\documentclass[runningheads]{llncs}

\usepackage{color,xcolor}
\usepackage{epsfig}
\usepackage{graphicx}

\usepackage{comment}

\usepackage{pifont}

\usepackage{microtype}
\usepackage[font=small]{caption}
\usepackage{arydshln}
\usepackage{tabularx}
\usepackage{adjustbox}
\usepackage{array}
\usepackage{booktabs}
\usepackage{colortbl}
\usepackage{float,wrapfig}
\usepackage{hhline}
\usepackage{multirow}
\usepackage{subcaption} %
\usepackage[percent]{overpic}

\usepackage{stmaryrd}

\usepackage{amsmath,amsfonts,amssymb}
\usepackage{bm}
\usepackage{nicefrac}
\usepackage{microtype}
\usepackage{dsfont}
\usepackage{changepage}
\usepackage{extramarks}
\usepackage{fancyhdr}
\usepackage{setspace}
\usepackage{soul}
\usepackage{xspace}

\usepackage[pagebackref=true,breaklinks=true,letterpaper=true,colorlinks,bookmarks=false]{hyperref}
\usepackage{url}

\usepackage{algorithm, algorithmic}
\usepackage{enumitem}
\usepackage[title]{appendix}

\usepackage[frozencache,cachedir=.]{minted}

\usepackage{amsmath,amsfonts,bm,relsize,dsfont}
\definecolor{MyDarkBlue}{rgb}{0,0.08,1}
\definecolor{MyDarkGreen}{rgb}{0.02,0.6,0.02}
\definecolor{MyDarkRed}{rgb}{0.8,0.02,0.02}
\definecolor{MyDarkOrange}{rgb}{0.40,0.2,0.02}
\definecolor{MyPurple}{RGB}{111,0,255}
\definecolor{MyRed}{rgb}{1.0,0.0,0.0}
\definecolor{MyGold}{rgb}{0.75,0.6,0.12}
\definecolor{MyDarkgray}{rgb}{0.66, 0.66, 0.66}
\definecolor{MyDarkCyan}{rgb}{0.05, 0.55, 0.45}
\definecolor{MyBlack}{rgb}{0., 0., 0.}
\definecolor{MyMagenta}{rgb}{1., 0., 1.}
\definecolor{BerkeleyYellow}{RGB}{255,204,41}
\definecolor{BerkeleyLightBlue}{RGB}{94,146,221}
\definecolor{BkDarkBlue}{rgb}{.05,.07,.353}

\newcommand{\red}[1]{\textcolor{MyRed}{#1}}
\newcommand{\cmark}{\ding{51}}%
\newcommand{\xmark}{\ding{55}}
\def\G{G}
\def\proj{H}
\def\Gdec{\G_{\text{dec}}}
\def\dim{K}
\def\Genc{\G_{\text{enc}}}
\def\Lgan{\mathcal{L}_\text{GAN}}
\def\Lpatchnce{\mathcal{L}_\text{PatchNCE}}
\def\Xset{X}
\def\Yset{Y}
\def\L{\mathcal{L}} %

\def\v{\bm{v}} %
\def\x{\bm{x}} %
\def\y{\bm{y}} %
\def\z{\bm{z}} %

\def\vpos{\bm{v^{+}}} %
\def\vneg{\bm{v^{-}}} %

\def\X{\mathcal{X}} %
\def\Y{\mathcal{Y}} %

\def\yhat{{\hat{\y}}}
\def\zhat{{\hat{\z}}}

\def\xtilde{{\tilde{\x}}}

\def\ztilde{{\tilde{\z}}}

\def\Re{\mathds{R}}
\newcommand{\expect}[1]{\mathbb{E}_{#1}}
\def\NCE{\ell}

\newcommand{\fid}{Fr\'echet Inception Distance\xspace}

\newcommand{\reffig}[1]{Figure~\ref{fig:#1}}
\newcommand{\refsec}[1]{Section~\ref{sec:#1}}
\newcommand{\refapp}[1]{Appendix~\ref{sec:#1}}
\newcommand{\reftbl}[1]{Table~\ref{tbl:#1}}

\newcommand{\refeq}[1]{Eqn.~\ref{eq:#1}}

\newcommand{\lblfig}[1]{\label{fig:#1}}
\newcommand{\lblsec}[1]{\label{sec:#1}}
\newcommand{\lbleq}[1]{\label{eq:#1}}
\newcommand{\lbltbl}[1]{\label{tbl:#1}}

\newcommand{\ignorethis}[1]{}

\newcommand{\myparagraph}[1]{\vspace{4pt} \noindent \textbf{#1}}

\def\1{\bm{1}}

\newcommand{\test}{\mathcal{D_{\mathrm{test}}}}

\newcolumntype{L}[1]{>{\raggedright\let\newline\\\arraybackslash\hspace{0pt}}m{#1}}
\newcolumntype{C}[1]{>{\centering\let\newline\\\arraybackslash\hspace{0pt}}m{#1}}
\newcolumntype{R}[1]{>{\raggedleft\let\newline\\\arraybackslash\hspace{0pt}}m{#1}}

\newcommand{\ignore}[1]{}

\makeatletter
\DeclareRobustCommand\onedot{\futurelet\@let@token\@onedot}
\def\@onedot{\ifx\@let@token.\else.\null\fi\xspace}
\def\eg{e.g\onedot,\xspace}

\makeatother

\begin{document}
\pagestyle{headings}
\mainmatter

\title{Contrastive Learning for Unpaired \\ Image-to-Image Translation}

\titlerunning{Contrastive Learning for Unpaired Image-to-Image Translation}
\authorrunning{Taesung Park, Alexei A. Efros, Richard Zhang, Jun-Yan Zhu}

\author{Taesung Park\inst{1} \hspace{1mm} Alexei A. Efros\inst{1} \hspace{1mm} Richard Zhang\inst{2} \hspace{1mm} Jun-Yan Zhu\inst{2}}

\institute{University of California, Berkeley\inst{1} 
\; \; 
Adobe Research\inst{2} \\
}

\maketitle

\begin{abstract}

In image-to-image translation, each patch in the output should reflect the {\em content} of the corresponding patch in the input, independent of domain. We propose a straightforward method for doing so -- maximizing mutual information between the two, using a framework based on contrastive learning. The method encourages two elements (corresponding patches) to map to a similar point in a learned feature space, relative to other elements (other patches) in the dataset, referred to as negatives. We explore several critical design choices for making contrastive learning effective in the image synthesis setting. Notably, we use a multilayer, patch-based approach, rather than operate on entire images. Furthermore, we draw negatives from \textit{within} the input image itself, rather than from the rest of the dataset. We demonstrate that our framework enables one-sided translation in the unpaired image-to-image translation setting, while improving quality and reducing training time. In addition, our method can even be extended to the training setting where each ``domain'' is only a single image.

\keywords{contrastive learning, noise contrastive estimation, mutual information, image generation}
\end{abstract}

\begin{figure}[t]
 \centering
 \includegraphics[width=1.\hsize]{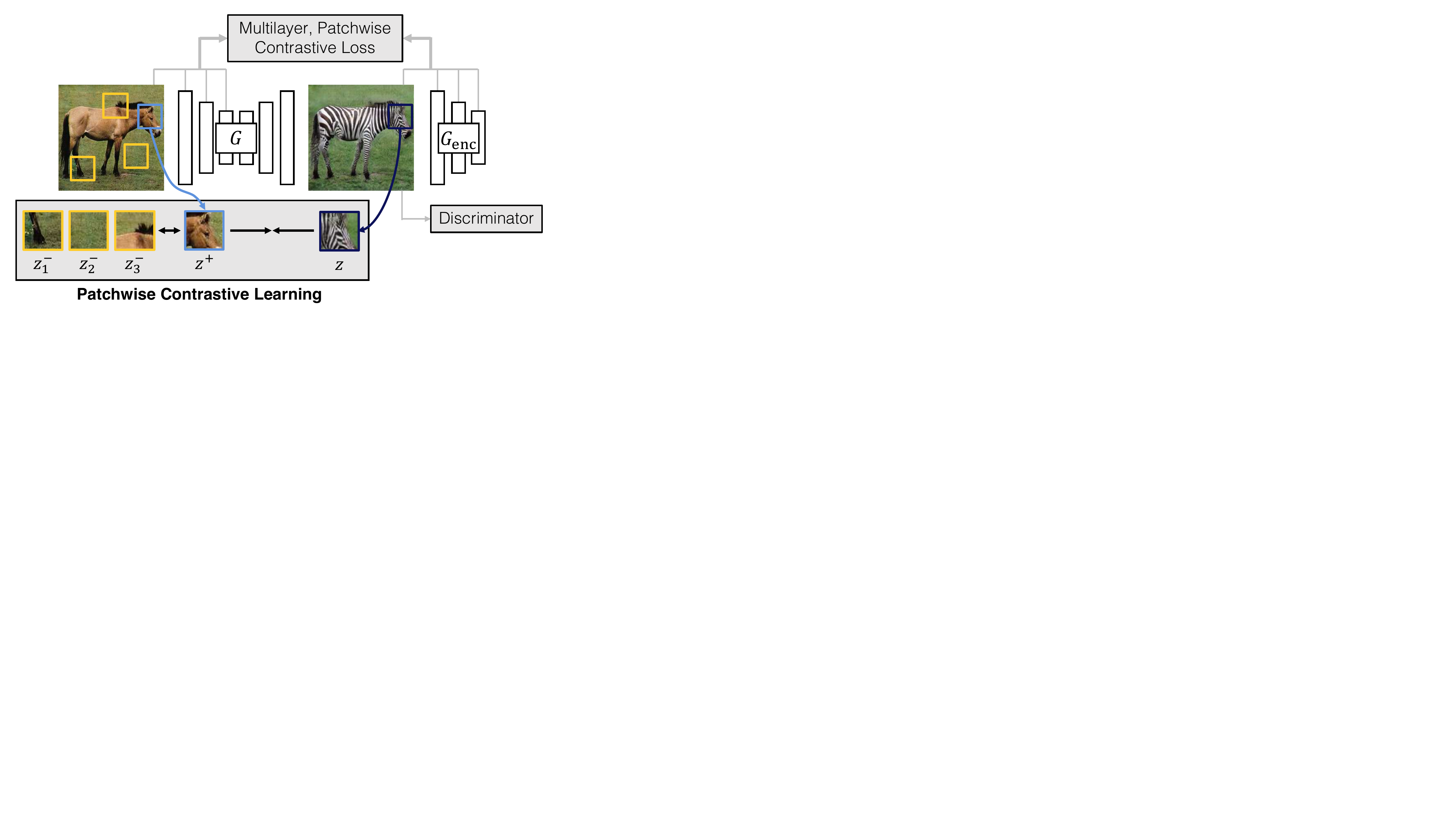}
  \vspace{-10pt}
 \caption{{\bf Patchwise Contrastive Learning for one-sided translation.} A generated \textcolor{BkDarkBlue}{\bf output patch} should appear closer to its \textcolor{BerkeleyLightBlue}{\bf corresponding input patch}, in comparison to other \textcolor{BerkeleyYellow}{\bf random patches}. We use a multilayer, patchwise contrastive loss, which maximizes \textit{mutual information} between corresponding input and output patches. This enables one-sided translation in the unpaired setting.}
 \lblfig{arch}
 \vspace{-10pt}
\end{figure}

\section{Introduction}

Consider the image-to-image translation problem in~\reffig{arch}. We wish for the output to take on the \textit{appearance} of the target domain (a zebra), while retaining the structure, or \textit{content}, of the specific input horse.  This is, fundamentally, a disentanglement problem: separating the content, which needs to be preserved across domains, from appearance, which must change.  
Typically, target appearance is enforced using an adversarial loss~\cite{goodfellow2014generative,isola2017image}, while content is preserved using cycle-consistency~\cite{zhu2017unpaired,yi2017dualgan,kim2017learning}. 
However, cycle-consistency assumes that the relationship between the two domains is a bijection, which is often too restrictive. %
In this paper, we propose an alternative, rather straightforward way of maintaining correspondence in content but not appearance -- by maximizing the mutual information between corresponding input and output patches.%

In a successful result, given a specific patch on the output, for example, the generated zebra forehead highlighted in blue, one should have a good idea that it came from the horse forehead, and not the other parts of the horse or the background vegetation. We achieve this by using a type of contrastive loss function,  InfoNCE loss~\cite{oord2018representation}, which aims to 
learn an embedding or an encoder that \textit{associates} corresponding patches to each other, while \textit{disassociating} them from others. To do so, the encoder learns to pay attention to the commonalities between the two domains, such as object parts and shapes, while being invariant to the differences, such as the textures of the animals.
The two networks, the generator and encoder, conspire together to generate an image such that patches can be easily traceable to the input.

Contrastive learning has been an effective tool in unsupervised visual representation learning~\cite{chen2020simple,he2019momentum,oord2018representation,wu2018unsupervised}.
In this work, we demonstrate its effectiveness in a conditional image synthesis setting and systematically study several key factors to make it successful. 
We find it pertinent to use it on a multilayer, patchwise fashion.
In addition, we find that drawing negatives \textit{internally} from within the input image, rather than externally from other images in the dataset, forces the patches to better preserve the content of the input. Our method requires neither memory bank~\cite{wu2018unsupervised,he2019momentum} nor specialized architectures~\cite{henaff2019data,bachman2019learning}.

Extensive experiments show that our faster, lighter model outperforms both prior one-sided translation methods~\cite{benaim2017one,fu2019geometry} and state-of-the-art models that rely on several auxiliary networks and multiple loss functions. Furthermore, since our contrastive representation is formulated within the same image, our method can even be trained on single images. Our code and models are available at \href{https://github.com/taesungp/contrastive-unpaired-translation}{GitHub}.

\section{Related Work}

\myparagraph{Image translation and cycle-consistency.}
Paired image-to-image translation~\cite{isola2017image} maps an image from input to output domain 
using an adversarial loss~\cite{goodfellow2014generative}, in conjunction with a reconstruction loss between the result and target. In unpaired translation settings, corresponding examples from domains are not available. In such cases, \textit{cycle-consistency} has become the de facto method for enforcing correspondence~\cite{zhu2017unpaired,yi2017dualgan,kim2017learning}, which learns an inverse mapping from the output domain back to the input and checks if the input can be reconstructed. Alternatively, UNIT~\cite{liu2017unsupervised} and MUNIT~\cite{huang2018multimodal} propose to learn a shared intermediate ``content'' latent space. Recent works further enable multiple domains and multi-modal synthesis~\cite{choi2017stargan,zhu2017toward,almahairi2018augmented,lee2018diverse,liu2019few} and improve the quality of results~\cite{tang2019attentiongan,zhang2019harmonic,gokaslan2018improving,wu2019transgaga,liang2018generative}.
In all of the above examples, cycle-consistency is used, often in multiple aspects, between (a) two image domains~\cite{kim2017learning,yi2017dualgan,zhu2017unpaired} (b) image to latent~\cite{choi2017stargan,huang2018multimodal,lee2018diverse,liu2017unsupervised,zhu2017toward}, or (c) latent to image~\cite{huang2018multimodal,zhu2017toward}. While effective, the underlying bijective assumption behind cycle-consistency is sometimes too restrictive. Perfect reconstruction is difficult to achieve, especially when images from one domain have additional information compared to the other domain.  %

\myparagraph{Relationship preservation.} An interesting alternative approach
is to encourage relationships present in the input be analogously reflected in the output. For example, perceptually similar patches \textit{within} an input image should be similar in the output~\cite{zhang2019harmonic}, output and input images share similar content regarding a pre-defined distance~\cite{bousmalis2017unsupervised,shrivastava2017learning,taigman2017unsupervised}, vector arithmetic between input images is preserved using a margin-based triplet loss~\cite{amodio2019travelgan},
distances \textit{between} input images should be consistent in output images~\cite{benaim2017one}, the network should be equivariant to geometric transformations~\cite{fu2019geometry}.
Among them, TraVeLGAN~\cite{amodio2019travelgan}, DistanceGAN~\cite{benaim2017one} and GcGAN~\cite{fu2019geometry} enable one-way translation and bypass cycle-consistency. However, they rely on relationship between entire images, or often with predefined distance functions. 
Here we seek to replace cycle-consistency by instead learning a cross-domain similarity function \textit{between input and output patches} through information maximization, without relying on a pre-specified distance.

\myparagraph{Emergent perceptual similarity in deep network embeddings.} Defining a ``perceptual'' distance function between high-dimensional signals, \eg images, has been a longstanding problem in computer vision and image processing. The majority of image translation work mentioned uses a per-pixel reconstruction metric, such as $\ell_1$. Such metrics do not reflect human perceptual preferences and can lead to blurry results. Recently, the deep learning community has found that the VGG classification network~\cite{simonyan2015very} trained on ImageNet dataset~\cite{deng2009imagenet} can be re-purposed as a ``perceptual loss''~\cite{dosovitskiy2016generating,gatys2016image,johnson2016perceptual,ulyanov2017improved,zhang2018unreasonable,mechrez2018contextual}, which can be used in paired image translation tasks~\cite{chen2017photographic,park2019semantic,wang2018pix2pixHD}, and was known to outperform traditional metrics such as SSIM~\cite{wang2004image} and FSIM~\cite{zhang2011fsim} on human perceptual tests~\cite{zhang2018unreasonable}. In particular, the Contextual Loss~\cite{mechrez2018contextual} boosts the perceptual quality of pretrained VGG features, validated by human perceptual judgments~\cite{mechrez2018maintaining}. 
In these cases, the frozen network weights cannot adapt to the data on hand. Furthermore, the frozen similarity function may not be appropriate when comparing data \textit{across} two domains, depending on the pairing.
By posing our constraint via mutual information, our method makes use of negative samples from the data, allowing the cross-domain similarity function to adapt to the particular input and output domains, and bypass using a pre-defined similarity function.

\myparagraph{Contrastive representation learning.} Traditional unsupervised learning has sought to learn a compressed code which can effectively reconstruct the input~\cite{hinton2006reducing}. Data imputation -- holding one subset of raw data to predict from another -- has emerged as a more effective family of pretext tasks, including denoising~\cite{vincent2008extracting}, context prediction~\cite{doersch2015unsupervised,pathak2016context}, colorization~\cite{larsson2017colorization,zhang2016colorful}, cross-channel encoding~\cite{zhang2017split}, frame prediction~\cite{lotter2016deep,misra2016shuffle}, and multi-sensory prediction~\cite{ngiam2011multimodal,owens2016ambient}. However, such methods suffer from the same issue as before --- the need for a pre-specified, hand-designed loss function to measure predictive performance.

Recently, a family of methods based on \textit{maximizing mutual information} has emerged to bypass the above issue~\cite{chen2020simple,he2019momentum,henaff2019data,hjelm2018learning,lowe2019putting,misra2019self,oord2018representation,tian2019contrastive,wu2018unsupervised}. These methods make use of noise contrastive estimation~\cite{gutmann2010noise}, learning an embedding where associated signals are brought together, in \textit{contrast} to other samples in the dataset (note that similar ideas go back to classic work on metric learning with Siamese nets~\cite{chopra2005learning}).   Associated signals can be an image with itself~\cite{malisiewicz-iccv11,shrivastava-sa11,dosovitskiy2015,he2019momentum,wu2018unsupervised}, an image with its downstream representation~\cite{hjelm2018learning,lowe2019putting}, neighboring patches within an image~\cite{isola2015,henaff2019data,oord2018representation}, or multiple views of the input image~\cite{tian2019contrastive}, and most successfully, an image with a set of transformed versions of itself~\cite{chen2020simple,misra2019self}. The design choices of the InfoNCE loss, such as the number of negatives and how to sample them, hyperparameter settings, and data augmentations all play a critical role and need to be carefully studied. We are the first to use InfoNCE loss for the conditional image synthesis tasks. As such, we draw on these important insights, and find additional pertinent factors, unique to image synthesis. %

\section{Methods}

We wish to translate images from input domain $\X \subset \Re^{H\times W\times C}$ to appear like an image from the output domain $\Y \subset \Re^{H\times W\times 3}$. We are given a dataset of unpaired instances $\Xset = \{\x\in \X\}, \Yset = \{\y\in \Y\}$. Our method can operate even when $\Xset$ and $\Yset$ only contain a single image each.

\begin{figure}[t]
 \centering
 \includegraphics[width=1.\hsize]{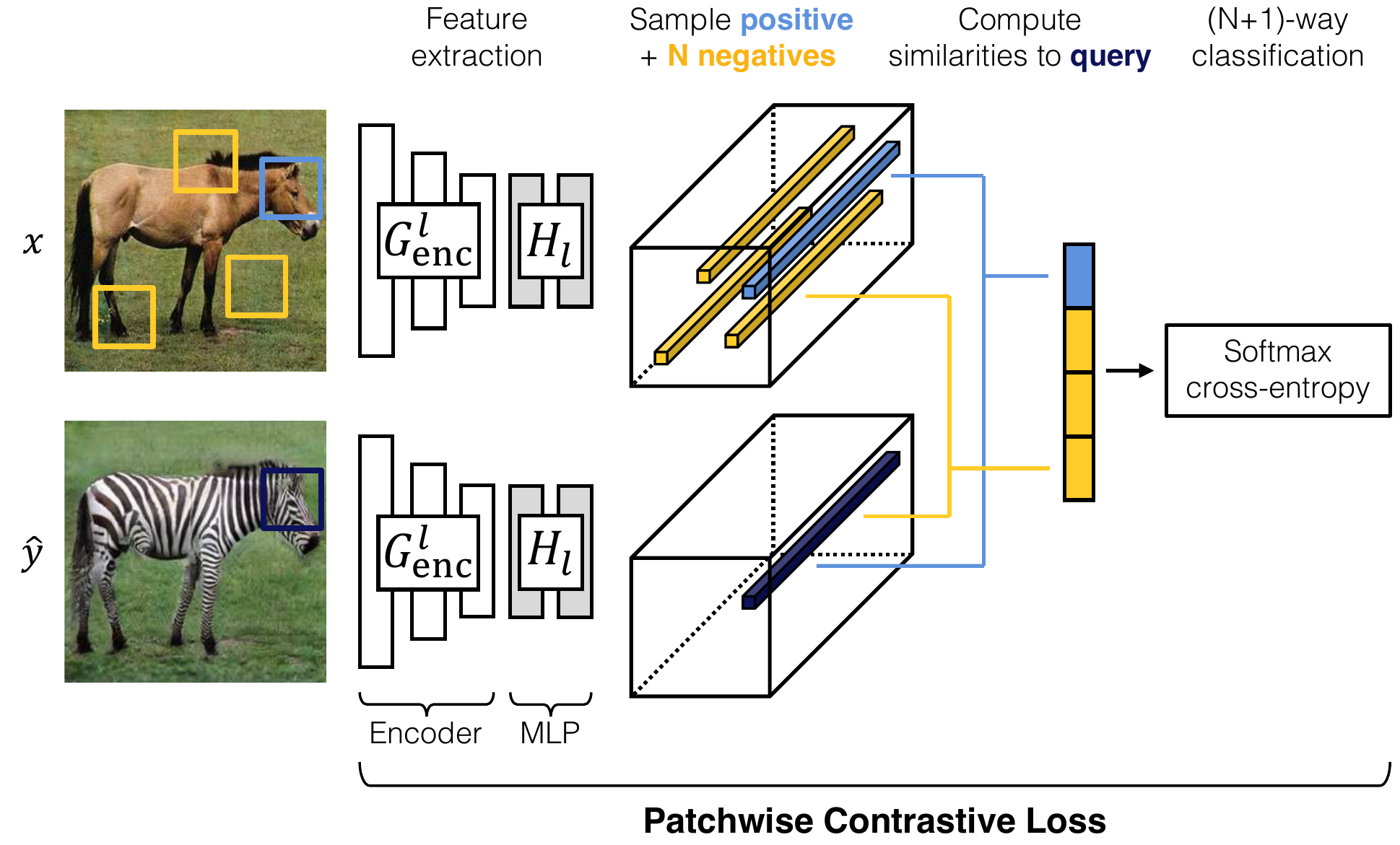}
\caption{{\bf Patchwise Contrastive Loss.} Both images, $\x$ and $\yhat$, are encoded into feature tensor. We sample a \textcolor{BkDarkBlue}{\bf query} patch from the output $\yhat$ and compare it to the input patch at the same location. We set up an (N+1)-way classification problem, where $N$ negative patches are sampled from the same input image at different locations. We reuse the encoder part $\Genc$ of our generator and add a two-layer MLP network. This network learns to project both the input and output patch to a shared embedding space.}
 \lblfig{nce}
 \vspace{-2mm}
\end{figure}

Our method only requires learning the mapping in one direction and avoids using inverse auxiliary generators and discriminators. This can largely simplify the training procedure and reduce training time. 
We break up our generator function $\G$ into two components, an encoder $\Genc$ followed by a decoder $\Gdec$, which are applied sequentially to produce output image $\yhat = G(\z) = \Gdec(\Genc(\x))$.

\myparagraph{Adversarial loss.} We use an adversarial loss~\cite{goodfellow2014generative}, to encourage the output to be visually similar to images from the target domain, as follows:

\begin{equation}
\Lgan(G, D, \Xset, \Yset) = \expect{\y\sim \Yset} \log D(\y) + \expect{\x \sim \Xset} \log(1-D(G(\x))).
\lbleq{gan}
\end{equation}

\myparagraph{Mutual information maximization.} We use a noise contrastive estimation framework~\cite{oord2018representation} to maximize mutual information between input and output. The idea of contrastive learning is to associate two signals, a ``query'' and its ``positive'' example, in contrast to other points within the dataset, referred to as ``negatives''. The query, positive, and $N$ negatives are mapped to $\dim$-dimensional vectors $\v, \vpos \in \Re^{\dim}$ and $\vneg\in \Re^{N\times \dim}$, respectively. $\vneg_n \in \Re^{\dim}$ denotes the n-th negative.
We normalize vectors onto a unit sphere to prevent the space from collapsing or expanding. An $(N+1)$--way classification problem is set up, where the distances between the query and other examples are scaled by a temperature $\tau=0.07$ and passed as logits~\cite{wu2018unsupervised,he2019momentum}. The cross-entropy loss is calculated, representing the probability of the positive example being selected over negatives.

\begin{equation}
\NCE(\v, \vpos, \vneg) = -\log{ \Bigg[
\frac{\exp (\v\cdot\vpos/\tau)}{\exp (\v\cdot\vpos/\tau) + \sum_{n=1}^N \exp (\v\cdot\vneg_n/\tau)} \Bigg] }.
\lbleq{nce_basics}
\end{equation}

\noindent Our goal is to associate the input and output data. In our context, query refers to an output. positive and negatives are corresponding and noncorresponding input. Below, we explore several important design choices, including how to map the images into vectors and how to sample the negatives.

\myparagraph{Multilayer, patchwise contrastive learning.} In the unsupervised learning setting, contrastive learning has been used both on an image and patch level~\cite{bachman2019learning,henaff2019data}.
For our application, we note that not only should the whole images share content, but also corresponding patches between the input and output images. For example, given a patch showing the legs of an output zebra, one should be able to more strongly associate it to the corresponding legs of the input horse, more so than the other patches of the horse image. Even at the pixel level, the colors of a zebra body (black and white) can be more strongly associated to the color of a horse body than to the background shades of grass. Thus, we employ a \textit{multilayer, patch-based} learning objective.

Since the encoder $\Genc$ is computed to produce the image translation, its feature stack is readily available, and we take advantage. Each layer and spatial location within this feature stack represents a patch of the input image, with deeper layers corresponding to bigger patches. We select $L$ layers of interest and pass the feature maps through a small two-layer MLP network $\proj_l$, as used in SimCLR~\cite{chen2020simple}, producing a stack of features $\{\z_l\}_L=\{\proj_l(\Genc^l(\x))\}_L$, where $\Genc^l$ represents the output of the $l$-th chosen layer. 
We index into layers $l\in \{1,2,...,L\}$ and denote $s\in \{1, ..., S_l\}$, where $S_l$ is the number of spatial locations in each layer.  %
We refer to the corresponding feature as $\z_l^s \in \Re^{C_l}$ and the other features as $\z_l^{S \setminus s}\in \Re^{(S_l-1)\times C_l}$, where $C_l$ is the number of channels at each layer. Similarly, we encode the output image $\yhat$ into $\{\zhat_l\}_L=\{\proj_l(\Genc^l(G(\x)))\}_L$.

We aim to match corresponding input-output patches at a specific location. We can leverage the other patches \textit{within} the input as negatives. For example, a zebra leg should be more closely associated with an input horse leg than the other patches of the same input, such as other horse parts or the background sky and vegetation. We name it as the \textit{PatchNCE} loss, as illustrated in \reffig{nce}. \refapp{app:code} provides pseudocode. 

\begin{equation}
\L_\text{PatchNCE}(G, \proj, \Xset) = \expect{\x \sim \Xset} \sum_{l=1}^L \sum_{s=1}^{S_l} \NCE(\zhat_l^s, \z_l^s, \z_l^{S\setminus s}).
\lbleq{nce_int}
\end{equation}

\noindent Alternatively, we can also leverage image patches from the rest of the dataset.
We encode a random negative image from the dataset $\xtilde$ into $\{\ztilde_l\}_L$, and use the following \textit{external} NCE loss. In this variant, we maintain a large, consistent dictionary of negatives using an auxiliary moving-averaged encoder, following MoCo~\cite{he2019momentum}. %
MoCo enables negatives to be sampled from a longer history, and performs more effective than end-to-end updates~\cite{oord2018representation,henaff2019data} and memory bank~\cite{wu2018unsupervised}. %
\begin{equation}
\L_\text{external}(G,\proj, \Xset)
= \expect{\x \sim \Xset, \ztilde \sim Z^-}\sum_{l=1}^L \sum_{s=1}^{S_l} \NCE(\zhat_l^s, \z_l^s, \ztilde_l),
\lbleq{nce_ext}
\end{equation}
where dataset negatives $\ztilde_l$ are sampled from an external dictionary $Z^-$ from the source domain, whose data are computed using a moving-averaged encoder $\hat{\proj}_l$ and moving-averaged MLP $\hat{\proj}$.  %
We refer our readers to the original work for more details~\cite{he2019momentum}. %

In \refsec{unpaired_results}, we show that our encoder $\Genc$  learns to capture domain-invariant  concepts, such as animal body, grass, and sky for horse $\rightarrow$ zebra, while our decoder $\Gdec$ learns to synthesize domain-specific features such as zebra stripes.  Interestingly, through systematic evaluations, we find that using internal patches only outperforms using external patches. We hypothesize that by using internal statistics, our encoder does not need to model large intra-class variation such as white horse vs. brown horse, which is not necessary for generating output zebras. Single image internal statistics has been proven effective in many vision tasks such as segmentation~\cite{isola2014crisp}, super-resolution, and denoising~\cite{zontak2011internal,shocher2018zero}.

\myparagraph{Final objective.} Our final objective is as follows. The generated image should be realistic, while patches in the input and output images should share correspondence.~\reffig{arch} illustrates our minimax learning objective. Additionally, we may utilize PatchNCE loss $\Lpatchnce(G, H, \Yset)$ on images from domain $\Y$ to prevent the generator from making unnecessary changes. This loss is essentially a learnable, domain-specific version of the identity loss, commonly used by previous unpaired translation methods~\cite{taigman2017unsupervised,zhu2017unpaired}. 

\begin{equation}
\Lgan(G,D,\Xset, \Yset) + \lambda_{\Xset}\Lpatchnce(G, \proj, \Xset) + \lambda_{\Yset}\Lpatchnce(G, \proj, \Yset).
\lbleq{full}
\end{equation}
\noindent We choose $\lambda_{\Xset}=1$ when we jointly train with the identity loss $\lambda_{\Yset}=1$, and choose a larger value $\lambda_{\Xset}=10$ without the identity loss ($\lambda_{\Yset}=0$) to compensate for the absence of the regularizer. We find that the former configuration, named \textit{Contrastive Unpaired Translation (CUT)} hereafter, achieves superior performance to existing methods, whereas the latter, named \textit{FastCUT}, can be thought as a faster and lighter version of CycleGAN. 
Our model is relatively simple compared to recent methods that often use 5-10 losses and hyper-parameters.

\myparagraph{Discussion. }
Li et al.~\cite{li2017alice} has shown that cycle-consistency loss is the upper bound of conditional entropy $\operatorname{H}(X|Y)$ (and $\operatorname{H}(Y|X)$). Therefore, minimizing cycle-consistency loss encourages the output $\yhat$ to be more dependent on input $\x$. This is related to our objective of maximizing the mutual information $\operatorname{I}(X, Y)$, as $\operatorname{I}(X, Y)=\operatorname{H}(X)-\operatorname{H}(X|Y)$. As entropy $\operatorname{H}(X)$ is a constant and independent of the generator $G$, maximizing mutual information is equivalent to minimizing the conditional entropy. Notably, using contrastive learning, we can achieve a similar goal without introducing inverse mapping networks and additional discriminators.  In the unconditional modeling scenario, InfoGAN~\cite{chen2018deeplab} shows that simple losses (e.g., L2 or cross-entropy) 
can serve as a lower bound for maximizing mutual information between an image and a low-dimensional code. In our setting, we maximize the mutual information between two high-dimensional image spaces, where simple losses are no longer effective.  
Liang et al.~\cite{liang2018generative} proposes an adversarial loss based on Siamese networks that encourages the output to be closer to the target domain than to its source domain. The above method still builds on cycle-consistency and two-way translations. Different from the above work, we use contrastive learning to enforce content consistency, rather than to improve the adversarial loss itself. To measure the similarity between two distributions, the Contextual Loss~\cite{mechrez2018contextual} used softmax over cosine disntances of features extracted from pre-trained networks. In contrast, we learn the encoder with the NCE loss to associate the input and output patches at the same location.

\begin{figure}[t]
 \centering
 \includegraphics[width=1.\hsize]{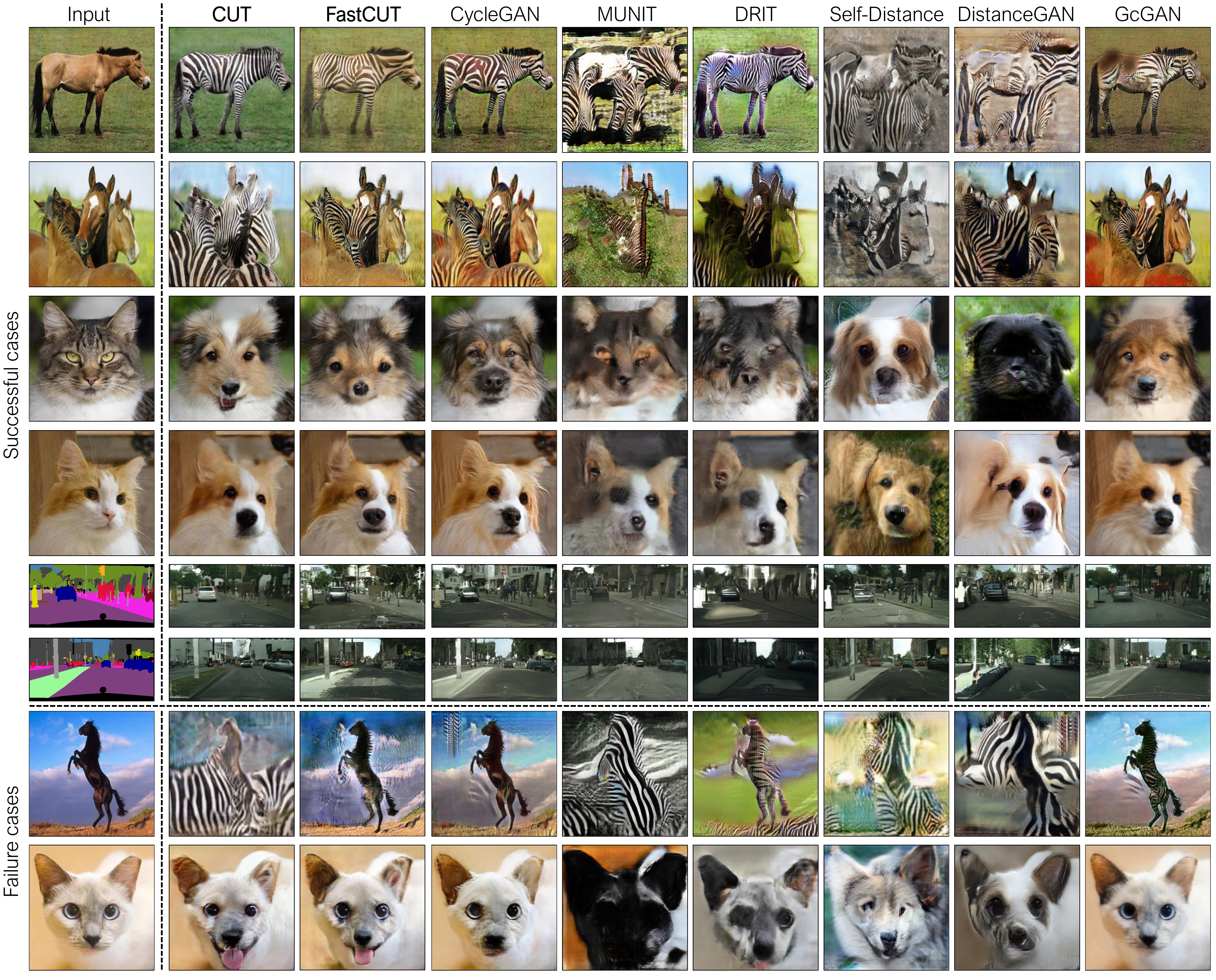} \vspace{-5mm}
\caption{\small {\bf Results}. We compare our methods (CUT and FastCUT) with existing methods on the horse$\rightarrow$zebra, cat$\rightarrow$dog, and Cityscapes datasets. CycleGAN~\cite{zhu2017unpaired}, MUNIT~\cite{liu2017unsupervised}, and DRIT~\cite{lee2018diverse}, are two-sided methods, while SelfDistance, DistanceGAN~\cite{benaim2017one}, and GcGAN~\cite{fu2019geometry} are one-sided. We show successful cases above the dotted lines. Our full version CUT is able to add the zebra texture to the horse bodies. Our fast variant FastCUT can also generate competitive results at the least computational cost of training. The final rows show failure cases. In the first, we are unable to identify the unfamiliar pose of the horse and instead add texture to the background. In the second, the method hallucinates a tongue.
}
\vspace{-2mm}
 \lblfig{baseline_comp}
 \vspace{-2mm}
\end{figure}

\section{Experiments}
\lblsec{expr}

We test across several datasets. We first show that our method improves upon baselines in unpaired image translation. We then show that our method can extend to \textit{single-image} training. Full results are available at our \href{https://taesungp.github.io/ContrastiveUnpairedTranslation}{website}. %

\begin{table}[t]

  {\small
    \centering
    \resizebox{1.0\linewidth}{!}{
    \begin{tabular}{l @{\hskip 2mm} c@{\hskip 2mm}ccc @{\hskip 3mm} c @{\hskip 3mm} c @{\hskip 2mm}cc}
    \toprule
    
    \multirow{2}{*}{\bf Method}  & \multicolumn{4}{c}{\bf Cityscapes} & {\bf Cat$\rightarrow$Dog} & \multicolumn{3}{c}{\bf Horse$\rightarrow$Zebra} \\ %
    \cmidrule(r){2-5} \cmidrule(r){6-6} \cmidrule(r){7-9}
    & {\bf mAP}$\uparrow$ & {\bf pixAcc}$\uparrow$ & {\bf classAcc}$\uparrow$ & {\bf FID}$\downarrow$ & {\bf FID}$\downarrow$ & {\bf FID}$\downarrow$ & {\bf sec/iter}$\downarrow$ & {\bf Mem(GB)}$\downarrow$ \\
        \midrule
    {CycleGAN}~\cite{zhu2017unpaired} & 20.4 & 55.9 & 25.4 & 76.3 & 85.9 & 77.2 & 0.40 & 4.81 \\ 
    {MUNIT}~\cite{liu2017unsupervised} & 16.9 & 56.5 & 22.5 & 91.4 & 104.4 & 133.8 & 0.39 & 3.84 \\
    {DRIT}~\cite{lee2018diverse} & 17.0 & 58.7 & 22.2 & 155.3 & 123.4 & 140.0 & 0.70 & 4.85 \\ \cdashline{1-9}
    {Distance}~\cite{benaim2017one} & 8.4 & 42.2 & 12.6 & 81.8 & 155.3 & 72.0 & {\bf 0.15} & 2.72  \\ 
    {SelfDistance}~\cite{benaim2017one} & 15.3 & 56.9 & 20.6 & 78.8 & 144.4 & 80.8 & 0.16 & 2.72  \\ 
    {GCGAN}~\cite{fu2019geometry} & 21.2 & 63.2 & 26.6 & 105.2 & 96.6 & 86.7 & 0.26 & 2.67   \\
    \cdashline{1-9}
    {CUT} & {\bf 24.7} & {\bf 68.8} & {\bf 30.7} & {\bf 56.4} & {\bf 76.2 } & {\bf 45.5} & 0.24 & 3.33  \\
    {FastCUT} & 19.1 & 59.9 & 24.3 & 68.8 & 94.0 & 73.4 & {\bf 0.15} & {\bf 2.25}\\
    \bottomrule
    \end{tabular}
    }

    \vspace{1mm}

    \caption{\small \textbf{Comparison with baselines}
    We compare our methods across datasets on common evaluation metrics. CUT denotes our model trained with the identity loss ($\lambda_{\Xset}=\lambda_{\Yset}=1$), and FastCUT without it ($\lambda_{\Xset}=10, \lambda_{\Yset}=0$).
    We show FID, a measure of image quality~\cite{heusel2017gans} (lower is better). For Cityscapes, we show the semantic segmentation scores (mAP, pixAcc, classAcc) to assess the discovered correspondence (higher is better for all metrics). Based on quantitative measures, CUT produces higher quality and more accurate generations with light footprint in terms of training speed (seconds per sample) and GPU memory usage. Our variant FastCUT also produces competitive results with even lighter computation cost of training.}
    \vspace{-5mm}
    \lbltbl{baselines}
    }
\end{table}

\myparagraph{Training details.}
\lblsec{details}
We follow the setting of CycleGAN~\cite{zhu2017unpaired}, except that the $\ell_1$ cycle-consistency loss is replaced with our contrastive loss. In detail, we used LSGAN~\cite{mao2017least} and Resnet-based generator~\cite{johnson2016perceptual} with PatchGAN~\cite{isola2017image}. We define our encoder as the first half of the generator, and accordingly extract our multilayer features from five evenly distributed points of the encoder. For single image translation, we use a StyelGAN2-based generator~\cite{karras2020analyzing}. To embed the encoder's features, we apply a two-layer MLP with 256 units at each layer. We normalize the vector by its L2 norm. See \refapp{app:training} for more training details.

\subsection{Unpaired image translation}
\lblsec{unpaired_results}

\myparagraph{Datasets} We conduct experiments on the following datasets.
\begin{itemize}[leftmargin=*,noitemsep,topsep=0pt,label=\textbullet]
\item \textit{Cat$\rightarrow$Dog} contains 5,000 training and 500 val images from AFHQ Dataset~\cite{choi2019stargan}.
\item \textit{Horse$\rightarrow$Zebra} contains 2,403 training and 260 zebra images from ImageNet~\cite{deng2009imagenet} and was introduced in CycleGAN~\cite{zhu2017unpaired}.
\item \textit{Cityscapes}~\cite{cordts2016cityscapes} contains street scenes from German cities, with 2,975 training and 500 validation images. We train models at 256$\times$256 resolution. Unlike previous datasets listed, this does have corresponding labels. We can leverage this to measure how well our unpaired algorithm discovers correspondences.
\end{itemize}

\myparagraph{Evaluation protocol.} We adopt the evaluation protocols from~\cite{heusel2017gans,zhu2017unpaired}, aimed at assessing \textit{visual quality} and \textit{discovered  correspondence}. For the first, we utilize the widely-used \fid (FID) metric, which empirically estimates the distribution of real and generated images in a deep network space and computes the divergence between them. Intuitively, if the generated images are realistic, they should have similar summary statistics as real images, in any feature space. For \textit{Cityscapes} specifically, we have ground truth of paired label maps. If accurate correspondences are discovered, the algorithm should generate images that are recognizable as the correct class. Using an off-the-shelf network to test ``semantic interpretability'' of image translation results has been commonly used~\cite{zhang2016colorful,isola2017image}. We use the pretrained semantic segmentation network DRN\cite{yu2017dilated}. We train the DRN at 256x128 resolution, and compute mean average precision (mAP), pixel-wise accuracy (pixAcc), and average class accuracy (classAcc). See \refapp{app:evaluation} for more evaluation details.

\myparagraph{Comparison to baselines.} In \reftbl{baselines}, we show quantitative measures of our and \reffig{baseline_comp}, we compare our method to baselines. We present two settings of our method in \refeq{full}: CUT with the identity loss ($\lambda_{\Xset}=\lambda_{\Yset}=1$), and FastCUT without it ($\lambda_{\Xset}=10,\lambda_{\Yset}=0$). On image quality metrics across datasets, our methods outperform baselines. We show qualitative results in \reffig{baseline_comp} and additional results in \refapp{app:results}. 
In addition, our Cityscapes semantic segmentation scores are higher, suggesting that our method is able to find correspondences between output and input. %

\myparagraph{Speed and memory.} Since our model is one-sided, our method is memory-efficient and fast. For example, our method with the identity loss was 40\% faster and 31\% more memory-efficient than CycleGAN at training time, using the same  architectures as CycleGAN (\reftbl{baselines}). Furthermore, our faster variant FastCUT is 63\% faster and 53\% lighter, while achieving superior metrics to CycleGAN. \reftbl{baselines} contains the speed and memory usage of each method measured on NVIDIA GTX 1080Ti, and shows that FastCUT achieves competitive FIDs and segmentation scores with a lower time and memory requirement. Therefore, our method can serves as a practical, lighter alternative in scenarios, when an image translation model is jointly trained with other components~\cite{hoffman2018cycada,rao2020rl}.

\begin{figure}[t]
  \centering
  \begin{minipage}[b]{0.61\linewidth}  
  \resizebox{1.\linewidth}{!}{
    
    \begin{tabular}{l @{\hskip 4mm} ccccc @{\hskip 4mm} c @{\hskip 2mm} cc}
    \toprule
    \multirow{3}{*}{\bf Method} & \multicolumn{5}{c}{\bf Training settings} & \multicolumn{3}{c}{\bf Testing datasets} \\ 
    \cmidrule(r){2-6} \cmidrule(r){7-9}
     & \multirow{2}{*}{\bf Id} & \multirow{2}{*}{\bf Negs} & \multirow{2}{*}{\bf Layers} & \multirow{2}{*}{\bf Int} & \multirow{2}{*}{\bf Ext} & \shortstack{{\bf Horse$\rightarrow$}\\{\bf Zebra}} & \multicolumn{2}{c}{\bf Cityscapes} \\ \cmidrule(r){7-7} \cmidrule(r){8-9}
    & & & & & & {\bf FID}$\shortdownarrow$ & {\bf FID}$\shortdownarrow$ & {\bf mAP}$\shortuparrow$ \\
    \midrule 

    {CUT (default)} & \cmark & 255 & All & \cmark & \xmark & 45.5 & {\bf 56.4} & {\bf 24.7} \\
    \cdashline{1-9}
    
    { \hspace{0.5mm} no id}  & \red{\xmark} & 255 & All & \cmark & \xmark & 39.3 & 68.5 & 22.0 \\
    { \hspace{0.5mm} no id, 15 neg}  & \red{\xmark} & \red{15} & All & \cmark & \xmark & 44.1 & 59.7 & 23.1 \\
    { \hspace{0.5mm} no id, 15 neg, last} & \red{\xmark} & \red{15} & \red{Last} & \cmark & \xmark & {\bf 38.1} & 114.1 & 16.0 \\
    \cdashline{1-9}
    {\hspace{1.5mm} last}  & \cmark & 255 & \red{Last} & \cmark & \xmark & 441.7 & 141.1 & 14.9 \\
    {\hspace{1.5mm} int and ext}  & \cmark & 255 & All & \cmark & \red{\cmark} & 56.4 & 64.4 & 20.0 \\
    {\hspace{1.5mm} ext only} & \cmark & 255 & All & \red{\xmark} & \red{\cmark} & 53.0 & 110.3 & 16.5 \\
    {\hspace{1.5mm} ext only, last} & \cmark & 255 & \red{Last} & \red{\xmark} & \red{\cmark} & 60.1 & 389.1 & 5.6 \\

    \bottomrule
    \end{tabular}

   } \\
	
  \end{minipage}
  \begin{minipage}[t]{0.32\linewidth}
  \includegraphics[width=\linewidth]{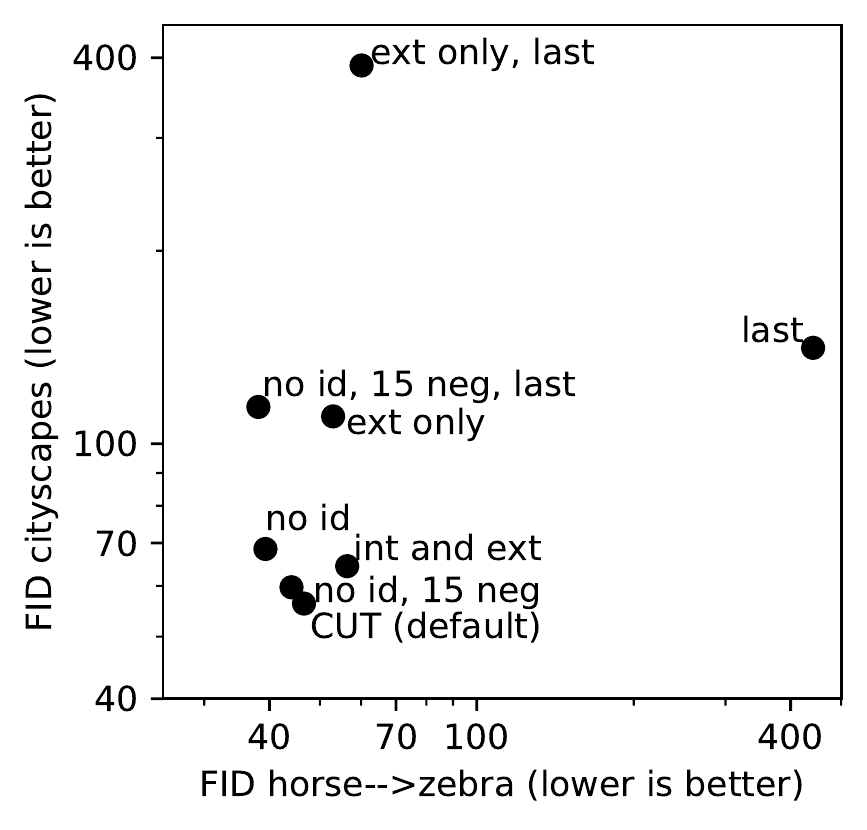} 
  \end{minipage}
  \vspace{-3mm}
  \caption{\small \textbf{Ablations.} The PatchNCE loss is trained with negatives from each layer output of the same (internal) image, with the identity preservation regularization. {\it (Left)} We try removing the identity loss [\textbf{Id}], using less negatives [\textbf{Negs}], using only the last layer of the encoder [\textbf{Layers}], and varying where patches are sampled, internal [{\bf Int}] vs external [{\bf Ext}].
  {\it (Right)} We plot the FIDs on horse$\rightarrow$zebra and Cityscapes dataset. Removing the identity loss (\texttt{no id}) and reducing negatives (\texttt{no id, 15 neg}) still perform strongly. In fact, our variant FastCUT does not use the identity loss. However, reducing number of layers (\texttt{last}) or using external patches (\texttt{ext}) hurts performance.
    }
  \vspace{-2mm}
  \lblfig{variants}
\end{figure}

\begin{figure}[h]
 \centering
 \includegraphics[width=1.\hsize]{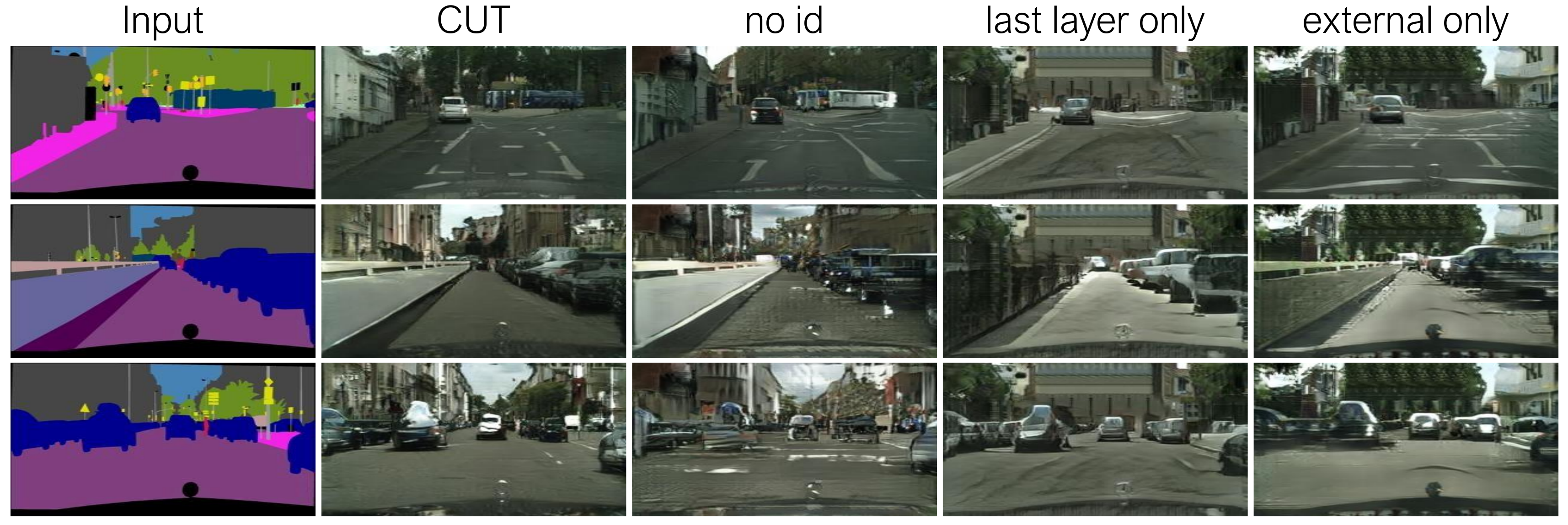} 
 \vspace{-7mm} 
 \caption{\small {\bf Qualitative ablation results} of our full method (CUT) are shown: {\em without} the identity loss $\Lpatchnce(G, H, \Yset)$ on domain $\Yset$ (\texttt{no id}), using only one layer of the encoder (\texttt{last layer only}), and using external instead of internal negatives (\texttt{external only}). The ablations cause noticeable drop in quality, including repeated building or vegetation textures when using only external negatives or the last layer output.
}
 \vspace{-3mm}
 \lblfig{ablations}
\end{figure}

\begin{figure}[h!]
 \centering
 \includegraphics[width=1.\hsize]{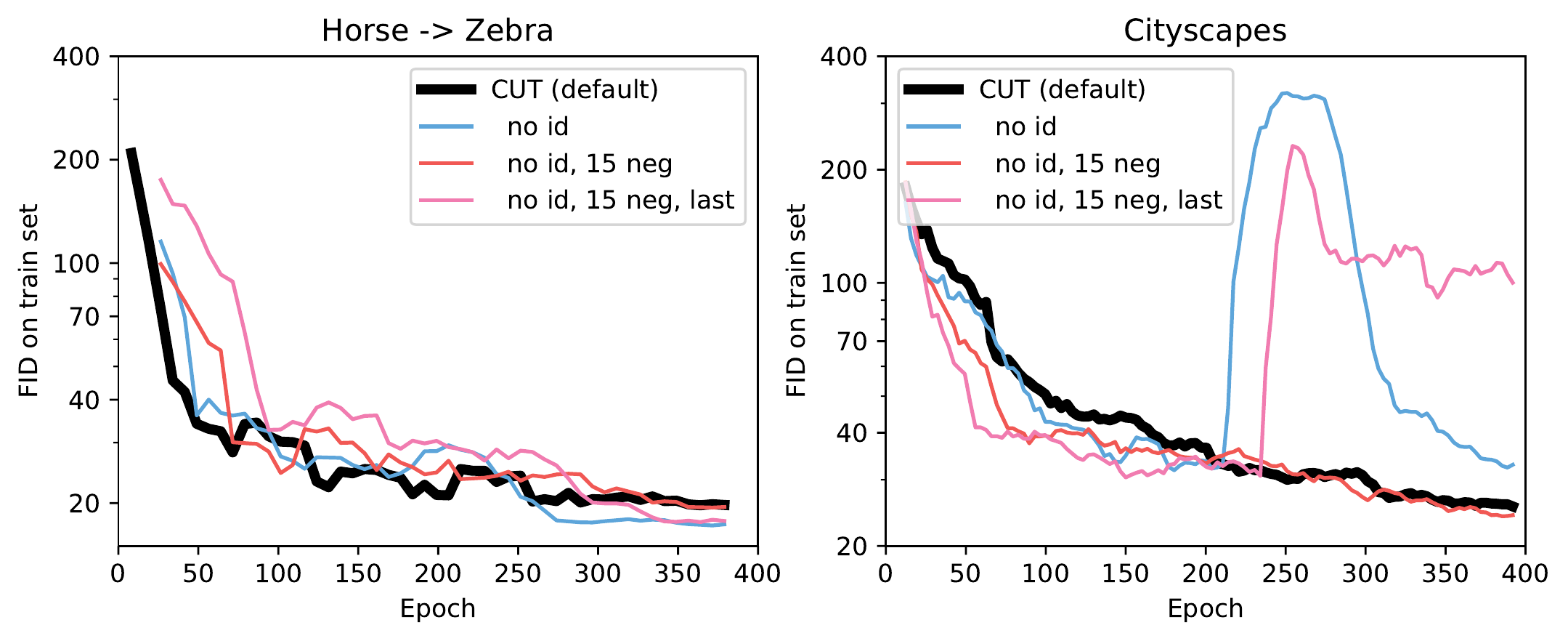}
\vspace{-5mm}
\caption{\small {\bf Identity loss $\Lpatchnce(G, H, \Yset)$ on domain $\Yset$ adds stability.} This regularizer encourages an image from the output domain $\y$ to be unchanged by the generator. Using it (shown in {\bf bold, black} curves), we observe better stability in comparison to other variants. On the left, our variant without the regularizer, \texttt{no id}, achieves better FID. However, we see higher variance in the training curve. On the right, training without the regularizer can lead to collapse.}
\vspace{-6mm}
 \lblfig{noidentity}
\end{figure}

\subsection{Ablation study and analysis}
\lblsec{analysis}
We find that in the image synthesis setting, similarly to the unsupervised learning setting~\cite{henaff2019data,he2019momentum,chen2020simple}, implementation choices for contrastive loss are important. Here, try various settings and ablations of our method, summarized in \reffig{variants}.
By default, we use the ResNet-based generator used in CycleGAN~\cite{zhu2017unpaired}, with patchNCE using (a) negatives sampled from the input image, (b) multiple layers of the encoder, and (c) a PatchNCE loss $\Lpatchnce(G, H, \Yset)$ on domain $\Yset$. In \reffig{variants}, we show results using several variants and ablations, taken after training for 400 epochs. We show qualitative examples in \reffig{ablations}.

\myparagraph{Internal negatives are more effective than external.} By default, we sample negatives from \textit{within} the same image (internal negatives). We also try adding negatives from other images, using a momentum encoder~\cite{he2019momentum}. However, the external negatives, either as addition (\texttt{int and ext}) or replacement of internal negatives (\texttt{ext only}), hurts performance. In \reffig{ablations}, we see a loss of quality, such as repeated texture in the Cityscapes dataset, indicating that sampling negatives from the same image serves as a stronger signal for preserving content.

\myparagraph{Importance of using multiple layers of encoder.} Our method uses multiple layers of the encoder, every four layers from pixels to the $16^\text{th}$ layer. This is consistent with the standard use of $\ell_1$+VGG loss, which uses layers from the pixel level up to a deep convolutional layer. On the other hand, many contrastive learning-based unsupervised learning papers map the whole image into a single representation. To emulate this, we try only using the last layer of the encoder (\texttt{last}), and try a variant using external negatives only (\texttt{ext only, last}). Performance is drastically reduced in both cases. In unsupervised representation learning, the input images are fixed. For our application, the loss is being used as a signal for synthesizing an image. As such, this indicates that the dense supervision provided by using multiple layers of the encoder is important when performing image synthesis.

\myparagraph{$\Lpatchnce(G, H, \Yset)$ regularizer stabilizes training.} Given an image from the output domain $\y\in\Y$, this regularizer encourages the generator to leave the image unchanged with our patch-based contrastive loss. We also experiment with a variant without this regularizer, \texttt{no id}. As shown in \reffig{variants}, removing the regularizer improves results for the horse$\rightarrow$zebra task, but decreases performance on Cityscapes. We further investigate by showing the training curves in \reffig{noidentity}, across 400 epochs. In the Cityscapes results, the training can collapse without the regularizer (although it can recover). We observe that although the final FID is sometimes better without, the training is more stable with the regularizer. %

\myparagraph{Visualizing learned similarity by encoder $\Genc$}
To further understand why our encoder network $\Genc$ has learned to perform horse$\rightarrow$ zebra task, we study the output space of the 1st residual block for both horse and zebra features. As shown in \reffig{vis}. Given an input and output image, we compute the distance between a query patch's feature vector $\v$ (highlighted as red or blue dot) to feature vectors $\vneg$ of all the patches in the input using $\exp (\v\cdot\vneg/\tau)$ (\refeq{nce_basics}). Additionally, we perform a PCA dimension reduction on feature vectors from both horse and zebra patches. In (d) and (e), we show the top three principal components, which looks similar before and after translation. This indicates that our encoder is able to bring the corresponding patches from two domains into a similar location in the feature embedding space.  

\begin{figure}[t!]
 \centering
 \includegraphics[width=1.\hsize]{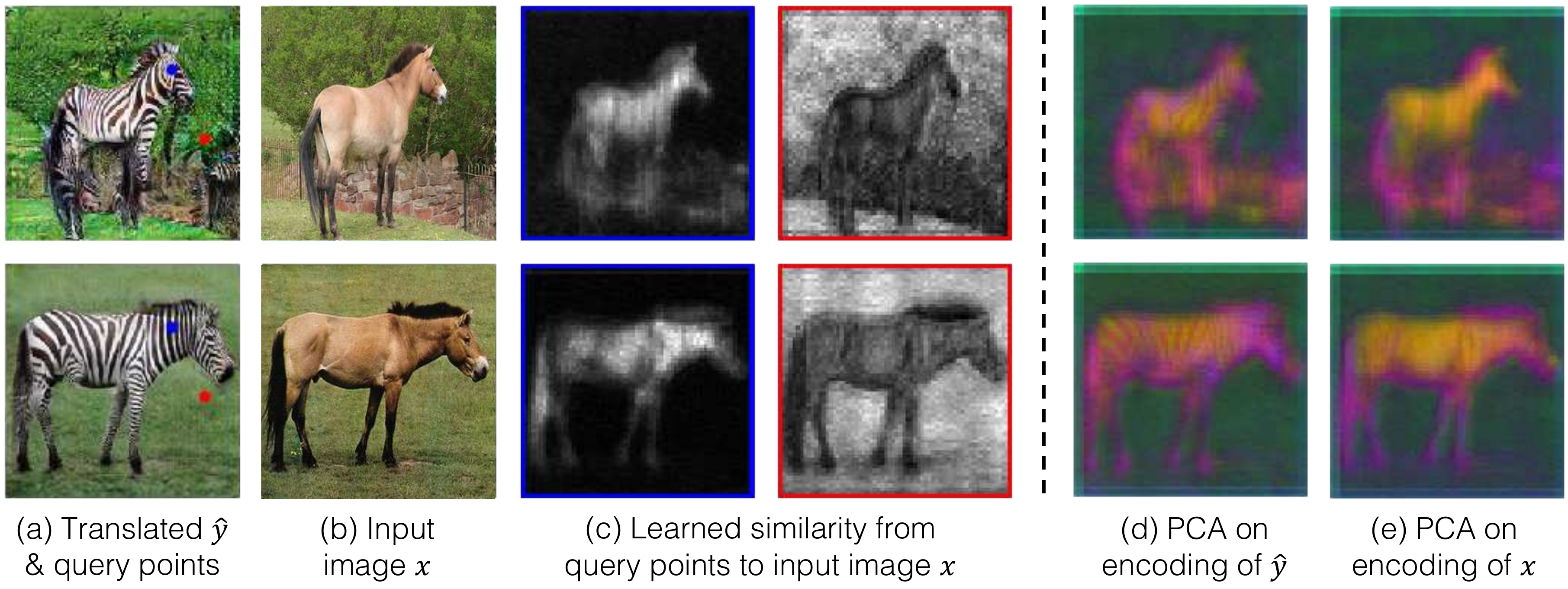}
  \vspace{-6mm}
 \caption{{\bf Visualizing the learned similarity by $\Genc$.}
  Given query points (blue or red) on an output image (a) and input (b), we visualize the learned similarity to patches on the input image by computing $\exp (\v\cdot\vneg/\tau)$ in (c). Here $\v$ is the query patch in the output and $\vneg$ denotes patches from the input. This suggests that our encoder may learn cross-domain correspondences implicitly. In (d) and (e), we visualize the top 3 PCA components of the shared embedding.
 }
 \lblfig{vis}
 \vspace{-5mm}
\end{figure}

\myparagraph{Additional applications}.  \reffig{app} shows additional results: Parisian street $\rightarrow$ Burano's brightly painted houses and Russian Blue cat $\rightarrow$ Grumpy cat.

\begin{figure}[h!]
 \centering
 \includegraphics[width=1.\hsize]{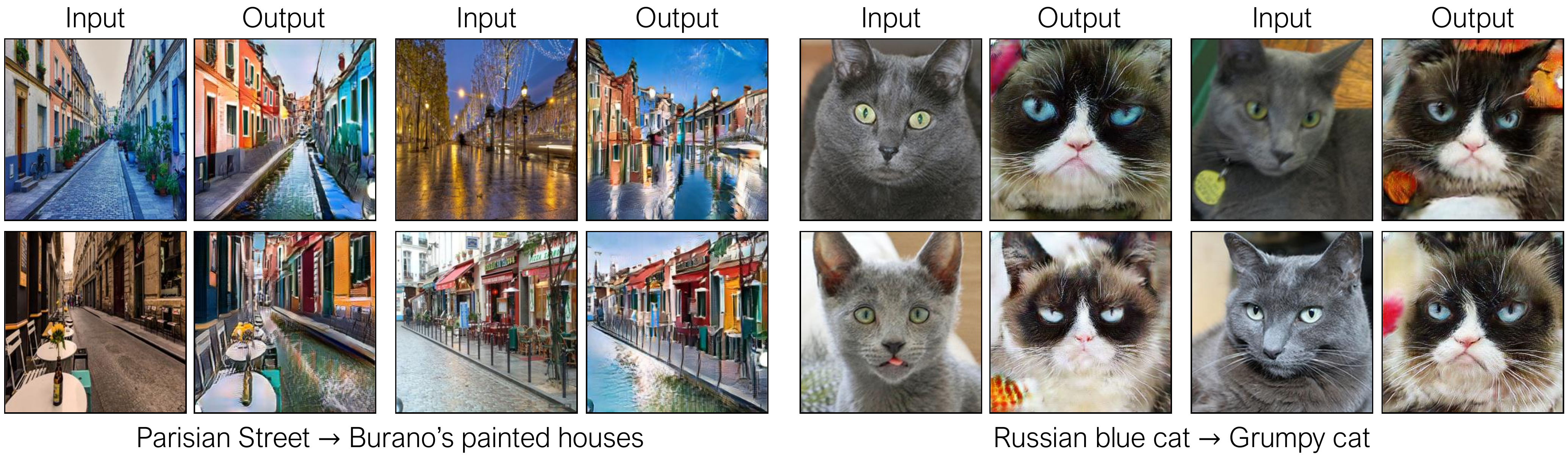}
  \vspace{-6mm}
 \caption{{\bf Additional applications} on Parisian street $\rightarrow$ Burano's colored houses and Russian Blue cat $\rightarrow$ Grumpy cat. }
 \lblfig{app}
 \vspace{-5mm}
\end{figure}

\subsection{High-resolution single image translation}

Finally, we conduct experiments in the single image setting, where both the source and target domain only have one image each. Here, we transfer a Claude Monet's painting to a natural photograph. Recent methods~\cite{shaham2019singan,shocher2018ingan} have explored training unconditional models on a single image. Bearing the additional challenge of respecting the structure of the input image, conditional image synthesis using only one image has not been explored by previous image-to-image translation methods. Our painting $\rightarrow$ photo task is also different from neural style transfer~\cite{gatys2016image,johnson2016perceptual} (photo $\rightarrow$ painting) and photo style transfer~\cite{luan2017deep,yoo2019photorealistic} (photo $\rightarrow$ photo).

Since the whole image (at HD resolution) cannot fit on a commercial GPU, at each iteration we train on 16 random crops of size 128$\times$128. We also randomly scale the image to prevent overfitting. Furthermore, we observe that limiting the receptive field of the discriminator is important for preserving the structure of the input image, as otherwise the GAN loss will force the output image to be identical to the target image. Therefore, the crops are further split into 64$\times$64 patches before passed to the discriminator. Lastly, we find that using gradient penalty~\cite{mescheder2018training,karras2019style} stabilizes optimization.   We call this variant SinCUT.   %

\reffig{single} shows a qualitative comparison between our results and baseline methods including two neural style transfer methods (Gatys et al.~\cite{gatys2016image} and STROTSS~\cite{kolkin2019style}), one leading photo style transfer method WCT$^2$~\cite{yoo2019photorealistic}, and a CycleGAN baseline~\cite{zhu2017unpaired} that uses the $\ell_1$ cycle-consistency loss instead of our contrastive loss at the patch level. The input paintings are high-res, ranging from 1k to 1.5k. \refapp{app:single} includes additional examples. 
We observe that Gatys et al.~\cite{gatys2016image} fails to synthesize realistic textures. Existing photo style transfer methods such as WCT$^2$ can only modify the color of the input image. Our method SinCUT outperforms CycleGAN and is comparable to a leading style transfer method~\cite{kolkin2019style}, which is based on optimal transport and self-similarity. Interestingly, our method is not originally designed for this application. This result suggests the intriguing connection between image-to-image translation and neural style transfer. %

\begin{figure}[t]
 \centering
  \includegraphics[width=1.\hsize]{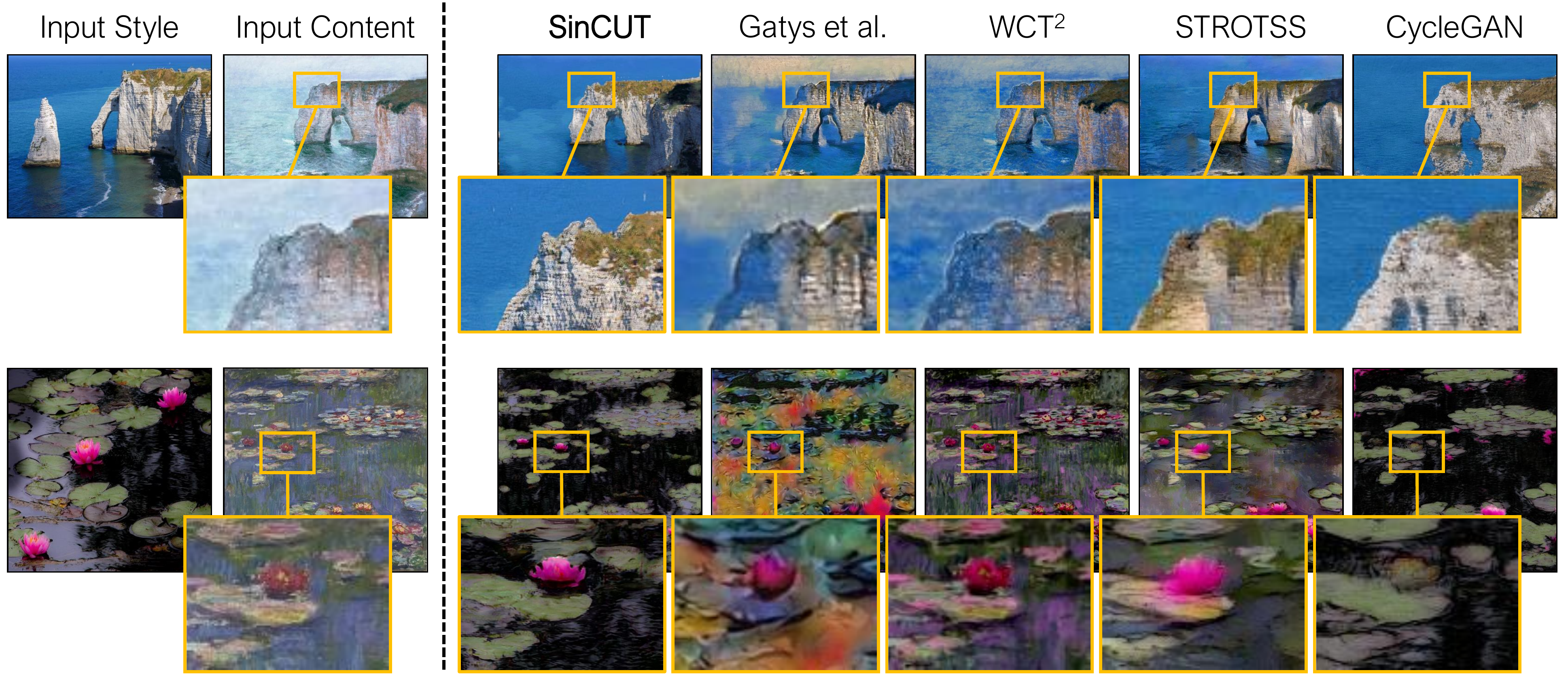}
  \vspace{-12pt}
 \caption{
 {\bf High-res painting to photo translation.}
 We transfer Claude Monet's paintings to reference natural photographs. The training only requires a single image from each domain.  We compare our results (SinCUT) to recent style and photo transfer methods including Gatys et al.~\cite{gatys2016image}, WCT$^2$~\cite{yoo2019photorealistic}, STROTSS~\cite{kolkin2019style}, and patch-based CycleGAN~\cite{zhu2017unpaired}. Our method generates can reproduce the texture of the reference photo while retaining structure of input painting. Our generation is at 1k $\sim$ 1.5k resolution.}
 \lblfig{single}
 \vspace{-13pt}
\end{figure}

\section{Conclusion}

We propose a straightforward method for encouraging content preservation in unpaired image translation problems -- by maximizing the mutual information between input and output with contrastive learning. The objective learns an embedding to bringing together corresponding patches in input and output, while pushing away noncorresponding ``negative'' patches. We study several important design choices. Interestingly, drawing negatives from \textit{within} the image itself, rather than other images, provides a stronger signal. Our method \textit{learns a cross-domain similarity function} and is the first image translation algorithm, to our knowledge, to not use any pre-defined similarity function (such as $\ell_1$ or perceptual loss). As our method does not rely on cycle-consistency, it can enable one-sided image translation, with better quality than established baselines. In addition, our method can be used for \textit{single-image} unpaired translation.

\myparagraph{Acknowledgments.} 
We thank Allan Jabri and Phillip Isola for helpful discussion and feedback. Taesung Park is supported by a Samsung Scholarship and an Adobe Research Fellowship, and some of this work was done as an Adobe Research intern. This work was partially supported by NSF grant IIS-1633310, grant from SAP, and gifts from Berkeley DeepDrive and Adobe.

\clearpage
\bibliographystyle{splncs04}
\bibliography{egbib}

\begin{thebibliography}{10}
\providecommand{\url}[1]{\texttt{#1}}
\providecommand{\urlprefix}{URL }
\providecommand{\doi}[1]{https://doi.org/#1}

\bibitem{almahairi2018augmented}
Almahairi, A., Rajeswar, S., Sordoni, A., Bachman, P., Courville, A.: Augmented
  cyclegan: Learning many-to-many mappings from unpaired data. In:
  International Conference on Machine Learning (ICML) (2018)

\bibitem{amodio2019travelgan}
Amodio, M., Krishnaswamy, S.: Travelgan: Image-to-image translation by
  transformation vector learning. In: Proceedings of the IEEE Conference on
  Computer Vision and Pattern Recognition. pp. 8983--8992 (2019)

\bibitem{bachman2019learning}
Bachman, P., Hjelm, R.D., Buchwalter, W.: Learning representations by
  maximizing mutual information across views. In: Advances in Neural
  Information Processing Systems (NeurIPS) (2019)

\bibitem{benaim2017one}
Benaim, S., Wolf, L.: One-sided unsupervised domain mapping. In: Advances in
  Neural Information Processing Systems (NeurIPS) (2017)

\bibitem{bousmalis2017unsupervised}
Bousmalis, K., Silberman, N., Dohan, D., Erhan, D., Krishnan, D.: Unsupervised
  pixel-level domain adaptation with generative adversarial networks. In: IEEE
  Conference on Computer Vision and Pattern Recognition (CVPR) (2017)

\bibitem{caesar2018coco}
Caesar, H., Uijlings, J., Ferrari, V.: Coco-stuff: Thing and stuff classes in
  context. In: IEEE Conference on Computer Vision and Pattern Recognition
  (CVPR) (2018)

\bibitem{chen2018deeplab}
Chen, L.C., Papandreou, G., Kokkinos, I., Murphy, K., Yuille, A.L.: Deeplab:
  Semantic image segmentation with deep convolutional nets, atrous convolution,
  and fully connected crfs. IEEE Transactions on Pattern Analysis and Machine
  Intelligence (TPAMI)  \textbf{40}(4),  834--848 (2018)

\bibitem{chen2017photographic}
Chen, Q., Koltun, V.: Photographic image synthesis with cascaded refinement
  networks. In: IEEE International Conference on Computer Vision (ICCV) (2017)

\bibitem{chen2020simple}
Chen, T., Kornblith, S., Norouzi, M., Hinton, G.: A simple framework for
  contrastive learning of visual representations. In: International Conference
  on Machine Learning (ICML) (2020)

\bibitem{choi2017stargan}
Choi, Y., Choi, M., Kim, M., Ha, J.W., Kim, S., Choo, J.: Stargan: Unified
  generative adversarial networks for multi-domain image-to-image translation.
  In: IEEE Conference on Computer Vision and Pattern Recognition (CVPR) (2018)

\bibitem{choi2019stargan}
Choi, Y., Uh, Y., Yoo, J., Ha, J.W.: Stargan v2: Diverse image synthesis for
  multiple domains. In: IEEE Conference on Computer Vision and Pattern
  Recognition (CVPR) (2020)

\bibitem{chopra2005learning}
Chopra, S., Hadsell, R., LeCun, Y.: Learning a similarity metric
  discriminatively, with application to face verification. In: IEEE Conference
  on Computer Vision and Pattern Recognition (CVPR) (2005)

\bibitem{cordts2016cityscapes}
Cordts, M., Omran, M., Ramos, S., Rehfeld, T., Enzweiler, M., Benenson, R.,
  Franke, U., Roth, S., Schiele, B.: The cityscapes dataset for semantic urban
  scene understanding. In: IEEE Conference on Computer Vision and Pattern
  Recognition (CVPR) (2016)

\bibitem{deng2009imagenet}
Deng, J., Dong, W., Socher, R., Li, L.J., Li, K., Fei-Fei, L.: {ImageNet: A
  Large-Scale Hierarchical Image Database}. In: IEEE Conference on Computer
  Vision and Pattern Recognition (CVPR) (2009)

\bibitem{doersch2015unsupervised}
Doersch, C., Gupta, A., Efros, A.A.: Unsupervised visual representation
  learning by context prediction. In: IEEE International Conference on Computer
  Vision (ICCV) (2015)

\bibitem{dosovitskiy2016generating}
Dosovitskiy, A., Brox, T.: Generating images with perceptual similarity metrics
  based on deep networks. In: Advances in Neural Information Processing Systems
  (2016)

\bibitem{dosovitskiy2015}
Dosovitskiy, A., Fischer, P., Springenberg, J.T., Riedmiller, M., Brox, T.:
  Discriminative unsupervised feature learning with exemplar convolutional
  neural networks. IEEE Transactions on Pattern Analysis and Machine
  Intelligence (TPAMI)  \textbf{38}(9),  1734--1747 (2015)

\bibitem{fu2019geometry}
Fu, H., Gong, M., Wang, C., Batmanghelich, K., Zhang, K., Tao, D.:
  Geometry-consistent generative adversarial networks for one-sided
  unsupervised domain mapping. In: IEEE Conference on Computer Vision and
  Pattern Recognition (CVPR) (2019)

\bibitem{gatys2016image}
Gatys, L.A., Ecker, A.S., Bethge, M.: Image style transfer using convolutional
  neural networks. In: IEEE Conference on Computer Vision and Pattern
  Recognition (CVPR) (2016)

\bibitem{gokaslan2018improving}
Gokaslan, A., Ramanujan, V., Ritchie, D., In~Kim, K., Tompkin, J.: Improving
  shape deformation in unsupervised image-to-image translation. In: European
  Conference on Computer Vision (ECCV) (2018)

\bibitem{goodfellow2014generative}
Goodfellow, I., Pouget-Abadie, J., Mirza, M., Xu, B., Warde-Farley, D., Ozair,
  S., Courville, A., Bengio, Y.: Generative adversarial nets. In: Advances in
  Neural Information Processing Systems (2014)

\bibitem{gu2018arbitrary}
Gu, S., Chen, C., Liao, J., Yuan, L.: Arbitrary style transfer with deep
  feature reshuffle. In: IEEE Conference on Computer Vision and Pattern
  Recognition (CVPR) (2018)

\bibitem{gutmann2010noise}
Gutmann, M., Hyv{\"a}rinen, A.: Noise-contrastive estimation: A new estimation
  principle for unnormalized statistical models. In: International Conference
  on Artificial Intelligence and Statistics (AISTATS) (2010)

\bibitem{he2019momentum}
He, K., Fan, H., Wu, Y., Xie, S., Girshick, R.: Momentum contrast for
  unsupervised visual representation learning. In: IEEE Conference on Computer
  Vision and Pattern Recognition (CVPR) (2020)

\bibitem{henaff2019data}
H{\'e}naff, O.J., Razavi, A., Doersch, C., Eslami, S., Oord, A.v.d.:
  Data-efficient image recognition with contrastive predictive coding. In: IEEE
  Conference on Computer Vision and Pattern Recognition (CVPR) (2019)

\bibitem{heusel2017gans}
Heusel, M., Ramsauer, H., Unterthiner, T., Nessler, B., Hochreiter, S.: {GANs}
  trained by a two time-scale update rule converge to a local {Nash}
  equilibrium. In: Advances in Neural Information Processing Systems (2017)

\bibitem{hinton2006reducing}
Hinton, G.E., Salakhutdinov, R.R.: Reducing the dimensionality of data with
  neural networks. Science  \textbf{313}(5786),  504--507 (2006)

\bibitem{hjelm2018learning}
Hjelm, R.D., Fedorov, A., Lavoie-Marchildon, S., Grewal, K., Bachman, P.,
  Trischler, A., Bengio, Y.: Learning deep representations by mutual
  information estimation and maximization. arXiv preprint arXiv:1808.06670
  (2018)

\bibitem{hoffman2018cycada}
Hoffman, J., Tzeng, E., Park, T., Zhu, J.Y., Isola, P., Saenko, K., Efros,
  A.A., Darrell, T.: Cycada: Cycle-consistent adversarial domain adaptation.
  In: International Conference on Machine Learning (ICML) (2018)

\bibitem{huang2018multimodal}
Huang, X., Liu, M.Y., Belongie, S., Kautz, J.: Multimodal unsupervised
  image-to-image translation. European Conference on Computer Vision (ECCV)
  (2018)

\bibitem{isola2017image}
Isola, P., Zhu, J.Y., Zhou, T., Efros, A.A.: Image-to-image translation with
  conditional adversarial networks. In: IEEE Conference on Computer Vision and
  Pattern Recognition (CVPR) (2017)

\bibitem{isola2014crisp}
Isola, P., Zoran, D., Krishnan, D., Adelson, E.H.: Crisp boundary detection
  using pointwise mutual information. In: European Conference on Computer
  Vision (ECCV) (2014)

\bibitem{isola2015}
Isola, P., Zoran, D., Krishnan, D., Adelson, E.H.: Learning visual groups from
  co-occurrences in space and time. arXiv preprint arXiv:1511.06811  (2015)

\bibitem{johnson2016perceptual}
Johnson, J., Alahi, A., Fei-Fei, L.: Perceptual losses for real-time style
  transfer and super-resolution. In: European Conference on Computer Vision
  (ECCV) (2016)

\bibitem{karras2019style}
Karras, T., Laine, S., Aila, T.: A style-based generator architecture for
  generative adversarial networks. In: IEEE Conference on Computer Vision and
  Pattern Recognition (CVPR) (2019)

\bibitem{karras2020analyzing}
Karras, T., Laine, S., Aittala, M., Hellsten, J., Lehtinen, J., Aila, T.:
  Analyzing and improving the image quality of stylegan. IEEE Conference on
  Computer Vision and Pattern Recognition (CVPR)  (2020)

\bibitem{kim2017learning}
Kim, T., Cha, M., Kim, H., Lee, J., Kim, J.: Learning to discover cross-domain
  relations with generative adversarial networks. In: International Conference
  on Machine Learning (ICML) (2017)

\bibitem{kingma2015adam}
Kingma, D.P., Ba, J.: Adam: A method for stochastic optimization. In:
  International Conference on Learning Representations (ICLR) (2015)

\bibitem{kolkin2019style}
Kolkin, N., Salavon, J., Shakhnarovich, G.: Style transfer by relaxed optimal
  transport and self-similarity. In: IEEE Conference on Computer Vision and
  Pattern Recognition (CVPR) (2019)

\bibitem{larsson2017colorization}
Larsson, G., Maire, M., Shakhnarovich, G.: Colorization as a proxy task for
  visual understanding. In: IEEE Conference on Computer Vision and Pattern
  Recognition (CVPR). pp. 6874--6883 (2017)

\bibitem{lee2018diverse}
Lee, H.Y., Tseng, H.Y., Huang, J.B., Singh, M.K., Yang, M.H.: Diverse
  image-to-image translation via disentangled representation. In: European
  Conference on Computer Vision (ECCV) (2018)

\bibitem{li2017alice}
Li, C., Liu, H., Chen, C., Pu, Y., Chen, L., Henao, R., Carin, L.: Alice:
  Towards understanding adversarial learning for joint distribution matching.
  In: Advances in Neural Information Processing Systems (2017)

\bibitem{liang2018generative}
Liang, X., Zhang, H., Lin, L., Xing, E.: Generative semantic manipulation with
  mask-contrasting gan. In: European Conference on Computer Vision (ECCV)
  (2018)

\bibitem{liu2017unsupervised}
Liu, M.Y., Breuel, T., Kautz, J.: Unsupervised image-to-image translation
  networks. In: Advances in Neural Information Processing Systems (2017)

\bibitem{liu2019few}
Liu, M.Y., Huang, X., Mallya, A., Karras, T., Aila, T., Lehtinen, J., Kautz,
  J.: Few-shot unsupervised image-to-image translation. In: IEEE International
  Conference on Computer Vision (ICCV) (2019)

\bibitem{lotter2016deep}
Lotter, W., Kreiman, G., Cox, D.: Deep predictive coding networks for video
  prediction and unsupervised learning. arXiv preprint arXiv:1605.08104  (2016)

\bibitem{lowe2019putting}
L{\"o}we, S., O'Connor, P., Veeling, B.: Putting an end to end-to-end:
  Gradient-isolated learning of representations. In: Advances in Neural
  Information Processing Systems (NeurIPS) (2019)

\bibitem{luan2017deep}
Luan, F., Paris, S., Shechtman, E., Bala, K.: Deep photo style transfer. In:
  IEEE Conference on Computer Vision and Pattern Recognition (CVPR) (2017)

\bibitem{malisiewicz-iccv11}
Malisiewicz, T., Gupta, A., Efros, A.A.: Ensemble of {E}xemplar-{SVM}s for
  object detection and beyond. In: IEEE International Conference on Computer
  Vision (ICCV) (2011)

\bibitem{mao2017least}
Mao, X., Li, Q., Xie, H., Lau, Y.R., Wang, Z., Smolley, S.P.: Least squares
  generative adversarial networks. In: IEEE International Conference on
  Computer Vision (ICCV) (2017)

\bibitem{mechrez2018maintaining}
Mechrez, R., Talmi, I., Shama, F., Zelnik-Manor, L.: Maintaining natural image
  statistics with the contextual loss. In: Asian Conference on Computer Vision
  (ACCV) (2018)

\bibitem{mechrez2018contextual}
Mechrez, R., Talmi, I., Zelnik-Manor, L.: The contextual loss for image
  transformation with non-aligned data. In: European Conference on Computer
  Vision (ECCV) (2018)

\bibitem{mescheder2018training}
Mescheder, L., Geiger, A., Nowozin, S.: Which training methods for gans do
  actually converge? In: International Conference on Machine Learning (ICML)
  (2018)

\bibitem{misra2019self}
Misra, I., van~der Maaten, L.: Self-supervised learning of pretext-invariant
  representations. arXiv preprint arXiv:1912.01991  (2019)

\bibitem{misra2016shuffle}
Misra, I., Zitnick, C.L., Hebert, M.: Shuffle and learn: unsupervised learning
  using temporal order verification. In: European Conference on Computer
  Vision. pp. 527--544. Springer (2016)

\bibitem{ngiam2011multimodal}
Ngiam, J., Khosla, A., Kim, M., Nam, J., Lee, H., Ng, A.Y.: Multimodal deep
  learning. In: International Conference on Machine Learning (ICML) (2011)

\bibitem{oord2018representation}
Oord, A.v.d., Li, Y., Vinyals, O.: Representation learning with contrastive
  predictive coding. arXiv preprint arXiv:1807.03748  (2018)

\bibitem{owens2016ambient}
Owens, A., Wu, J., McDermott, J.H., Freeman, W.T., Torralba, A.: Ambient sound
  provides supervision for visual learning. In: European Conference on Computer
  Vision (ECCV) (2016)

\bibitem{park2019semantic}
Park, T., Liu, M.Y., Wang, T.C., Zhu, J.Y.: Semantic image synthesis with
  spatially-adaptive normalization. In: IEEE Conference on Computer Vision and
  Pattern Recognition (CVPR) (2019)

\bibitem{pathak2016context}
Pathak, D., Krahenbuhl, P., Donahue, J., Darrell, T., Efros, A.A.: Context
  encoders: Feature learning by inpainting. In: Proceedings of the IEEE
  conference on computer vision and pattern recognition. pp. 2536--2544 (2016)

\bibitem{radford2016unsupervised}
Radford, A., Metz, L., Chintala, S.: Unsupervised representation learning with
  deep convolutional generative adversarial networks. In: International
  Conference on Learning Representations (ICLR) (2016)

\bibitem{rao2020rl}
Rao, K., Harris, C., Irpan, A., Levine, S., Ibarz, J., Khansari, M.:
  Rl-cyclegan: Reinforcement learning aware simulation-to-real. In: IEEE
  Conference on Computer Vision and Pattern Recognition (CVPR) (2020)

\bibitem{richter2016playing}
Richter, S.R., Vineet, V., Roth, S., Koltun, V.: Playing for data: Ground truth
  from computer games. In: European Conference on Computer Vision (ECCV) (2016)

\bibitem{shaham2019singan}
Shaham, T.R., Dekel, T., Michaeli, T.: Singan: Learning a generative model from
  a single natural image. In: IEEE International Conference on Computer Vision
  (ICCV) (2019)

\bibitem{shocher2018ingan}
Shocher, A., Bagon, S., Isola, P., Irani, M.: Ingan: Capturing and remapping
  the" dna" of a natural image. In: IEEE International Conference on Computer
  Vision (ICCV) (2019)

\bibitem{shocher2018zero}
Shocher, A., Cohen, N., Irani, M.: “zero-shot” super-resolution using deep
  internal learning. In: IEEE Conference on Computer Vision and Pattern
  Recognition (CVPR) (2018)

\bibitem{shrivastava-sa11}
Shrivastava, A., Malisiewicz, T., Gupta, A., Efros, A.A.: Data-driven visual
  similarity for cross-domain image matching. ACM Transactions on Graphics
  (SIGGRAPH Asia)  \textbf{30}(6) (2011)

\bibitem{shrivastava2017learning}
Shrivastava, A., Pfister, T., Tuzel, O., Susskind, J., Wang, W., Webb, R.:
  Learning from simulated and unsupervised images through adversarial training.
  In: IEEE Conference on Computer Vision and Pattern Recognition (CVPR) (2017)

\bibitem{simonyan2015very}
Simonyan, K., Zisserman, A.: Very deep convolutional networks for large-scale
  image recognition. In: International Conference on Learning Representations
  (ICLR) (2015)

\bibitem{szegedy2016rethinking}
Szegedy, C., Vanhoucke, V., Ioffe, S., Shlens, J., Wojna, Z.: Rethinking the
  inception architecture for computer vision. In: IEEE Conference on Computer
  Vision and Pattern Recognition (CVPR) (2016)

\bibitem{taigman2017unsupervised}
Taigman, Y., Polyak, A., Wolf, L.: Unsupervised cross-domain image generation.
  In: International Conference on Learning Representations (ICLR) (2017)

\bibitem{tang2019attentiongan}
Tang, H., Xu, D., Sebe, N., Yan, Y.: Attention-guided generative adversarial
  networks for unsupervised image-to-image translation. In: International Joint
  Conference on Neural Networks (IJCNN) (2019)

\bibitem{tian2019contrastive}
Tian, Y., Krishnan, D., Isola, P.: Contrastive multiview coding. arXiv preprint
  arXiv:1906.05849  (2019)

\bibitem{torralba2011unbiased}
Torralba, A., Efros, A.A.: Unbiased look at dataset bias. In: IEEE Conference
  on Computer Vision and Pattern Recognition (CVPR) (2011)

\bibitem{ulyanov2017improved}
Ulyanov, D., Vedaldi, A., Lempitsky, V.: Improved texture networks: Maximizing
  quality and diversity in feed-forward stylization and texture synthesis. In:
  IEEE Conference on Computer Vision and Pattern Recognition (CVPR) (2017)

\bibitem{vincent2008extracting}
Vincent, P., Larochelle, H., Bengio, Y., Manzagol, P.A.: Extracting and
  composing robust features with denoising autoencoders. In: International
  Conference on Machine Learning (ICML) (2008)

\bibitem{wang2018pix2pixHD}
Wang, T.C., Liu, M.Y., Zhu, J.Y., Tao, A., Kautz, J., Catanzaro, B.:
  High-resolution image synthesis and semantic manipulation with conditional
  gans. In: IEEE Conference on Computer Vision and Pattern Recognition (CVPR)
  (2018)

\bibitem{wang2004image}
Wang, Z., Bovik, A.C., Sheikh, H.R., Simoncelli, E.P.: Image quality
  assessment: from error visibility to structural similarity. IEEE transactions
  on image processing  \textbf{13}(4),  600--612 (2004)

\bibitem{wu2019transgaga}
Wu, W., Cao, K., Li, C., Qian, C., Loy, C.C.: Transgaga: Geometry-aware
  unsupervised image-to-image translation. In: IEEE Conference on Computer
  Vision and Pattern Recognition (CVPR) (2019)

\bibitem{wu2018unsupervised}
Wu, Z., Xiong, Y., Yu, S.X., Lin, D.: Unsupervised feature learning via
  non-parametric instance discrimination. In: IEEE Conference on Computer
  Vision and Pattern Recognition (CVPR) (2018)

\bibitem{yi2017dualgan}
Yi, Z., Zhang, H., Tan, P., Gong, M.: Dualgan: Unsupervised dual learning for
  image-to-image translation. In: IEEE International Conference on Computer
  Vision (ICCV) (2017)

\bibitem{yoo2019photorealistic}
Yoo, J., Uh, Y., Chun, S., Kang, B., Ha, J.W.: Photorealistic style transfer
  via wavelet transforms. In: IEEE International Conference on Computer Vision
  (ICCV) (2019)

\bibitem{yu2017dilated}
Yu, F., Koltun, V., Funkhouser, T.: Dilated residual networks. In: IEEE
  Conference on Computer Vision and Pattern Recognition (CVPR) (2017)

\bibitem{zhang2011fsim}
Zhang, L., Zhang, L., Mou, X., Zhang, D.: Fsim: A feature similarity index for
  image quality assessment. IEEE transactions on Image Processing
  \textbf{20}(8),  2378--2386 (2011)

\bibitem{zhang2016colorful}
Zhang, R., Isola, P., Efros, A.A.: Colorful image colorization. In: European
  Conference on Computer Vision (ECCV) (2016)

\bibitem{zhang2017split}
Zhang, R., Isola, P., Efros, A.A.: Split-brain autoencoders: Unsupervised
  learning by cross-channel prediction. In: IEEE Conference on Computer Vision
  and Pattern Recognition (CVPR) (2017)

\bibitem{zhang2018unreasonable}
Zhang, R., Isola, P., Efros, A.A., Shechtman, E., Wang, O.: The unreasonable
  effectiveness of deep features as a perceptual metric. In: IEEE Conference on
  Computer Vision and Pattern Recognition (CVPR). pp. 586--595 (2018)

\bibitem{zhang2019harmonic}
Zhang, R., Pfister, T., Li, J.: Harmonic unpaired image-to-image translation.
  In: International Conference on Learning Representations (ICLR) (2019)

\bibitem{zhu2017unpaired}
Zhu, J.Y., Park, T., Isola, P., Efros, A.A.: Unpaired image-to-image
  translation using cycle-consistent adversarial networks. In: IEEE
  International Conference on Computer Vision (ICCV) (2017)

\bibitem{zhu2017toward}
Zhu, J.Y., Zhang, R., Pathak, D., Darrell, T., Efros, A.A., Wang, O.,
  Shechtman, E.: Toward multimodal image-to-image translation. In: Advances in
  Neural Information Processing Systems (2017)

\bibitem{zontak2011internal}
Zontak, M., Irani, M.: Internal statistics of a single natural image. In: IEEE
  Conference on Computer Vision and Pattern Recognition (CVPR) (2011)

\end{thebibliography}

\begin{subappendices}
\renewcommand{\thesection}{\Alph{section}}%

\section{Additional Image-to-Image Results}
\lblsec{app:results}

We first show additional, randomly selected results on datasets used in our main paper. We then show results on additional datasets.

\subsection{Additional comparisons}

In \reffig{horse2zebra}, we show additional, randomly selected results for Horse$\rightarrow$Zebra and Cat$\rightarrow$Dog. This is an extension of Figure 3 in the main paper. We compare to baseline methods CycleGAN~\cite{zhu2017unpaired}, MUNIT~\cite{huang2018multimodal}, DRIT~\cite{lee2018diverse}, Self-Distance and DistanceGAN~\cite{benaim2017one}, and GcGAN~\cite{fu2019geometry}.

\subsection{Additional datasets}

In \reffig{appleyosemite} and \reffig{gta2cityscapes}, we show additional datasets, compared against baseline method CycleGAN~\cite{zhu2017unpaired}. Our method provides better or comparable results, demonstrating its flexibility across a variety of datasets.
\begin{itemize}[leftmargin=*,noitemsep,topsep=0pt,label=\textbullet]
\item \textit{Apple$\rightarrow$Orange} contains 996 apple and 1,020 orange images from ImageNet and was introduced in CycleGAN~\cite{zhu2017unpaired}.
\item \textit{Yosemite Summer$\rightarrow$Winter} contains 1,273 summer and 854 winter images of Yosemite scraped using the FlickAPI was introduced in CycleGAN~\cite{zhu2017unpaired}.
\item \textit{GTA$\rightarrow$Cityscapes} GTA contains 24,966 images~\cite{richter2016playing} and Cityscapes~\cite{cordts2016cityscapes} contains 19,998 images of street scenes from German cities. The task was originally used in CyCADA~\cite{hoffman2018cycada}.
\end{itemize}

\clearpage

\begin{figure}[h!]
 \centering
 \includegraphics[width=1.\linewidth]{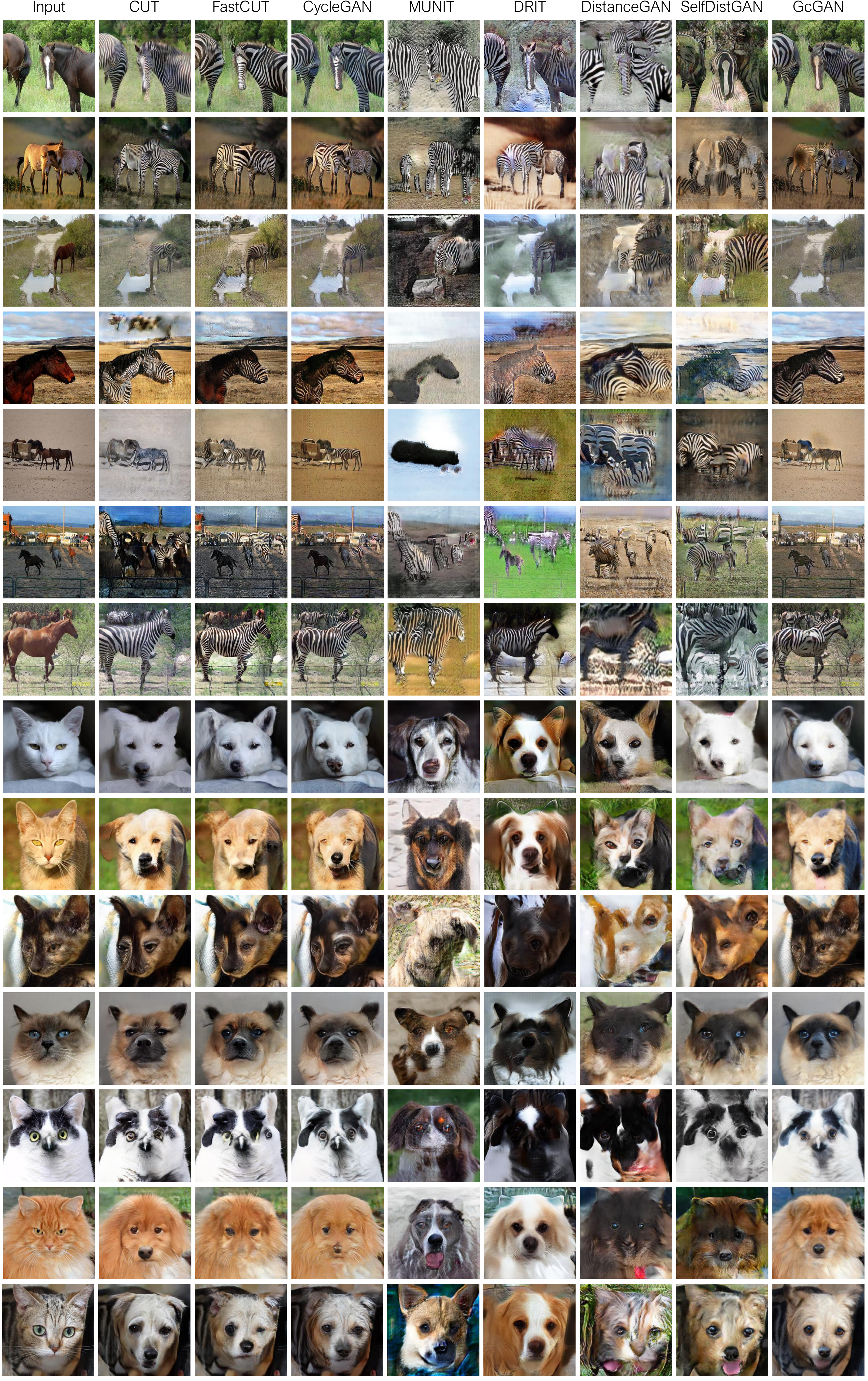} \vspace{-7mm}
\caption{\small {\bf Randomly selected Horse$\rightarrow$Zebra and Cat$\rightarrow$Dog results}. This is an extension of Figure 3 in the main paper.
}
\vspace{-5mm}
 \lblfig{horse2zebra}
 \vspace{-2mm}
\end{figure}

\begin{figure}[h]
 \centering
\includegraphics[width=1.\linewidth]{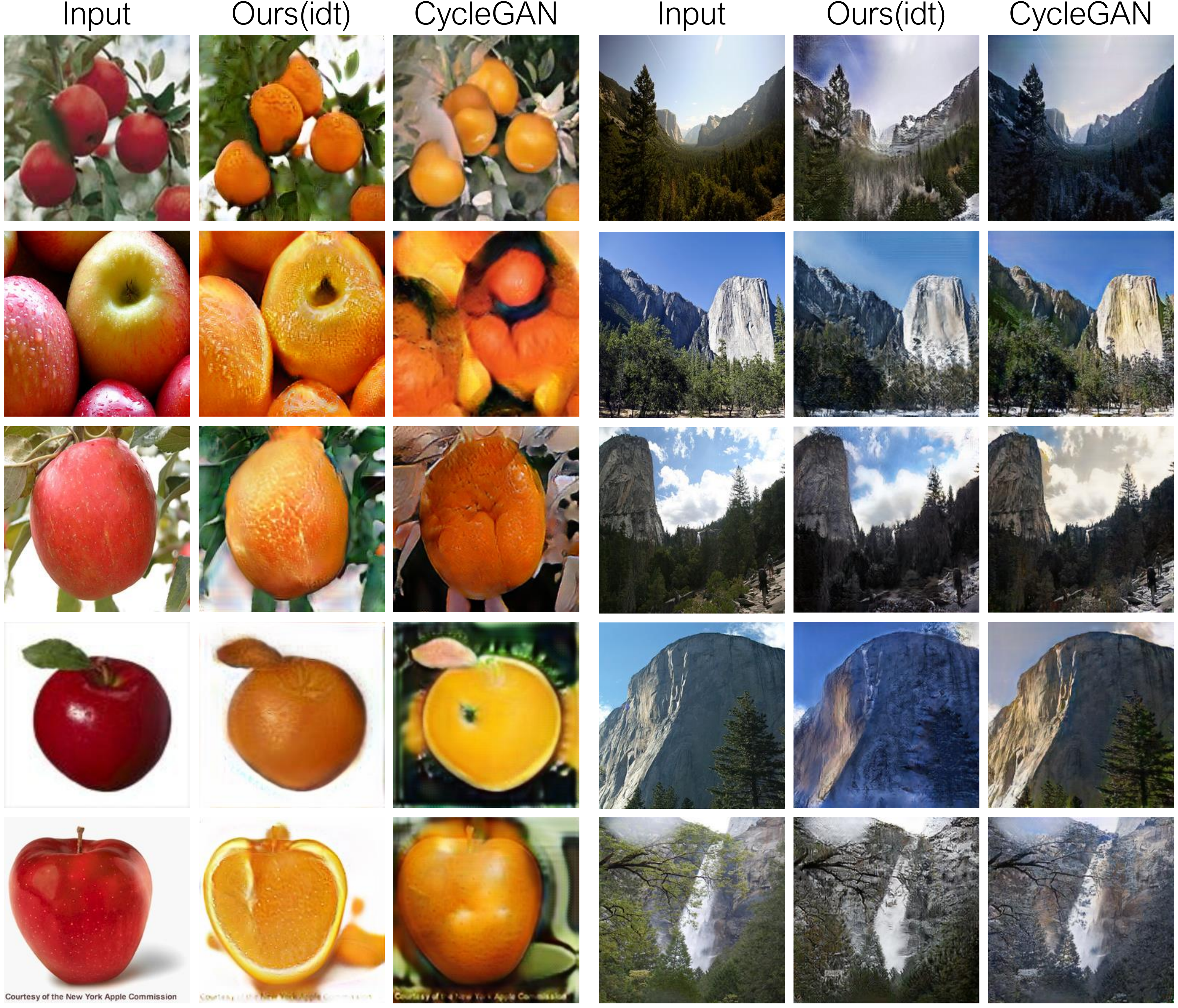} \vspace{-6mm}
\caption{\small {\bf Apple$\rightarrow$Orange} and {\bf Summer$\rightarrow$Winter Yosemite}. CycleGAN models were downloaded from the authors' public code repository. Apple$\rightarrow$Orange shows that CycleGAN may suffer from color flipping issue. 
}
\vspace{-2mm}
 \lblfig{appleyosemite}
 \vspace{-0mm}
\end{figure}

\begin{figure}[h]
 \centering
 \includegraphics[width=1.\hsize]{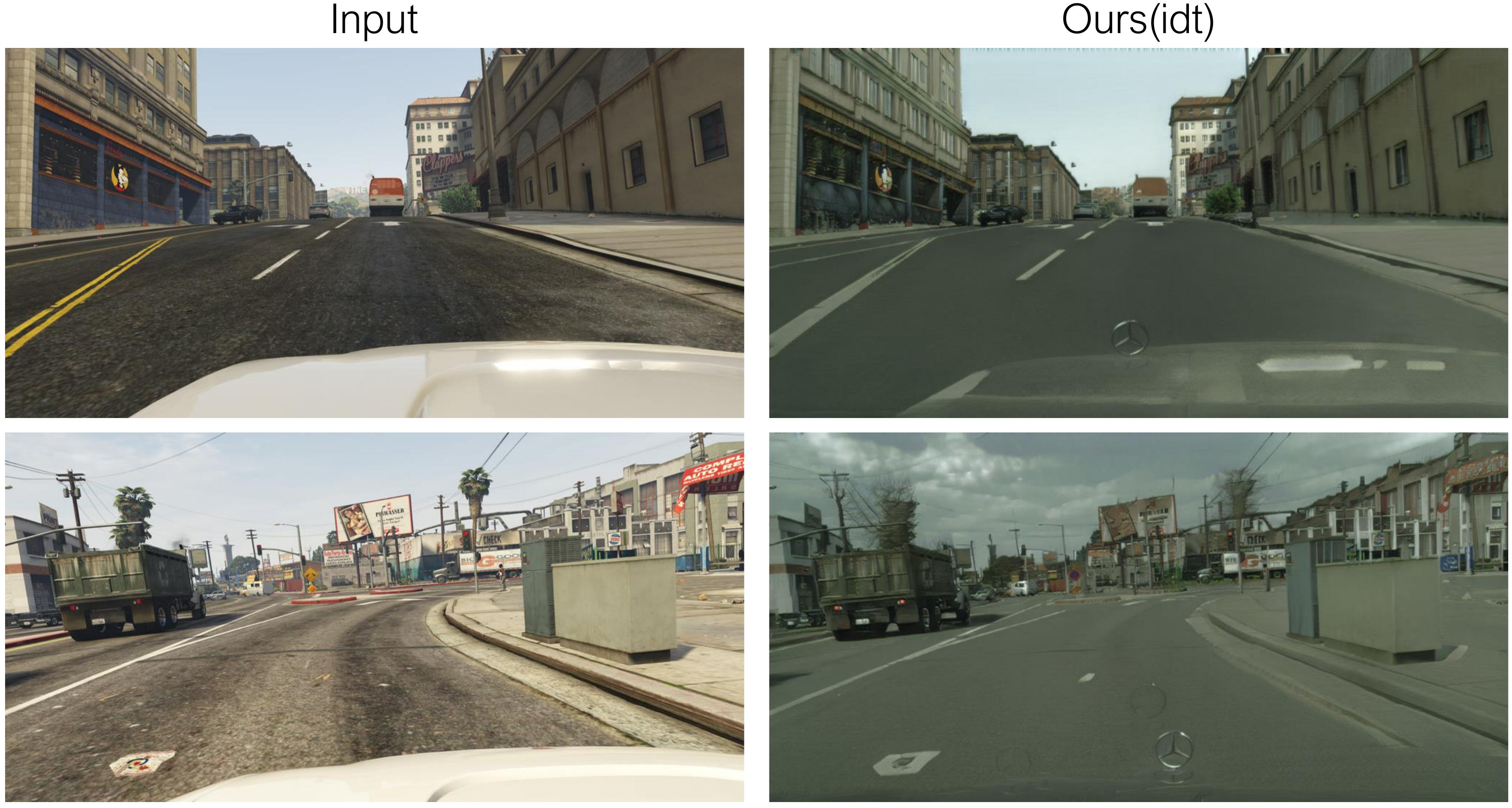} \vspace{-6mm}
\caption{\small {\bf GTA$\rightarrow$Cityscapes} results at $1024\times512$ resolution. The model was trained on $512\times512$ crops. }.

\vspace{-5mm}
 \lblfig{gta2cityscapes}
 \vspace{-2mm}
\end{figure}

\clearpage

\section{Additional Single Image Translation Results}
\lblsec{app:single}
We show additional results in \reffig{single1} and \reffig{single2}, 
and describe training details below.

\myparagraph{Training details.} At each iteration, the input image is randomly scaled to a width between 384 to 1024, and we randomly sample 16 crops of size $128\times 128$. To avoid overfitting, we divide crops into $64\times64$ tiles before passing them to the discriminator. At test time, since the generator network is fully convolutional, it takes the input image at full size. 

We found that adopting the architecture of StyleGAN2~\cite{karras2020analyzing} instead of CycleGAN slightly improves the output quality, although the difference is marginal. Our StyleGAN2-based generator consists of one downsampling block of StyleGAN2 discriminator, 6 StyleGAN2 residual blocks, and one StyleGAN2 upsampling block. Our discriminator has the same architecture as StyleGAN2. Following StyleGAN2, we use non-saturating GAN loss~\cite{radford2016unsupervised} with R1 gradient penalty~\cite{mescheder2018training}. Since we do not use style code, the style modulation layer of StyleGAN2 was removed.  

\myparagraph{Single image results.} 

In \reffig{single1} and \reffig{single2}, 
we show additional comparison results for our method,  Gatys et al.~\cite{gatys2016image}, STROTSS~\cite{kolkin2019style}, WCT$^2$~\cite{yoo2019photorealistic}, and CycleGAN baseline~\cite{zhu2017unpaired}. Note that the CycleGAN baseline adopts the same augmentation techniques as well as the same generator/discriminator architectures as our method. The image resolution is at 1-2 Megapixels. Please zoom in to see more visual details. 

Both figures demonstrate that our results look more photorealistic compared to CycleGAN baseline, Gatys et al~\cite{gatys2016image}, and WCT$^2$. The quality of our results is on par with results from STROTSS~\cite{kolkin2019style}. Note that STROTSS~\cite{kolkin2019style} compares to and outperforms recent style transfer methods (e.g., ~\cite{gu2018arbitrary,mechrez2018contextual}).

\begin{figure}[t]
 \centering
  \includegraphics[angle=90, width=0.92\hsize]{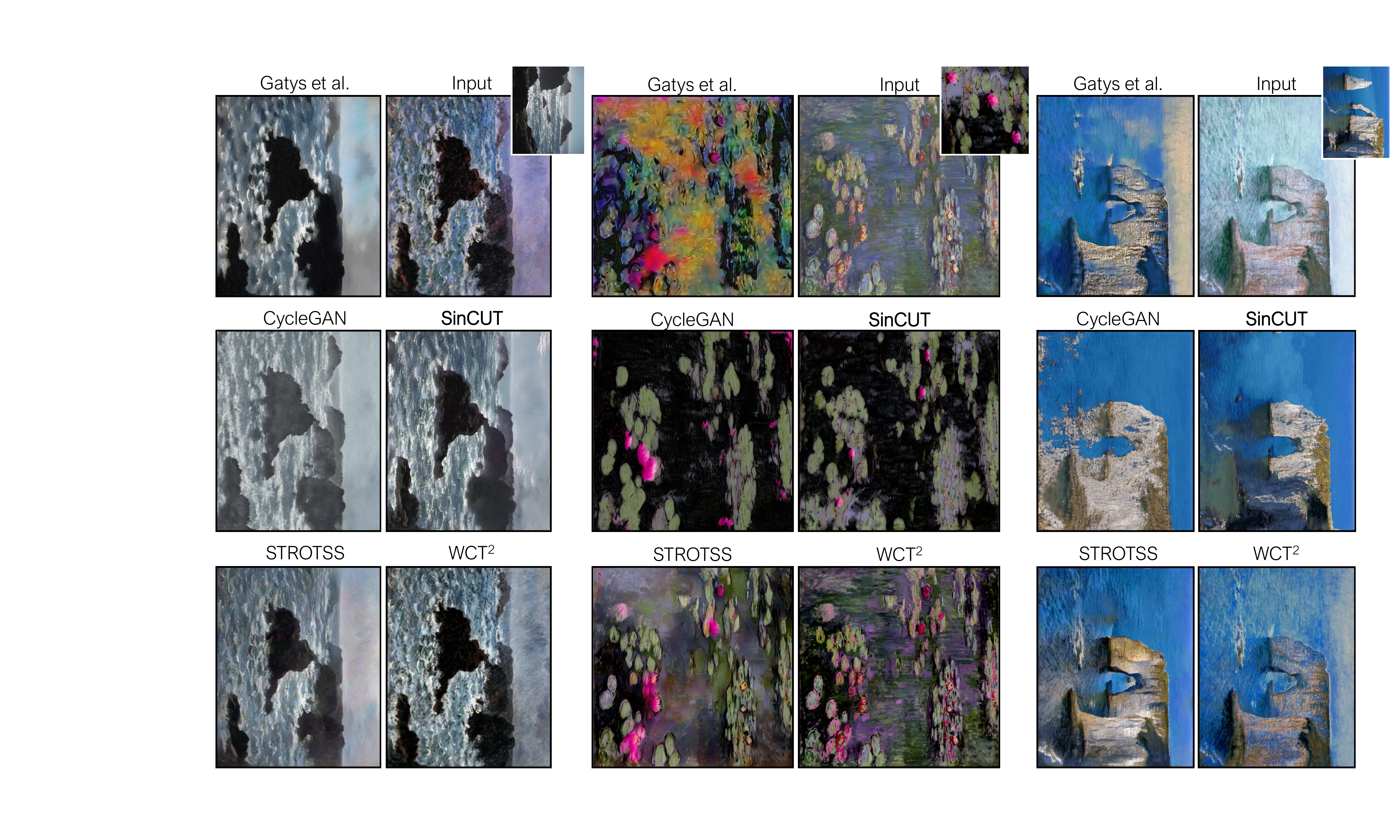}
  \vspace{-3mm}
 \caption{{\bf High-res painting to photo translation (I).} We transfer Monet's paintings to reference natural photos shown as insets at top-left corners. The training only requires a single image from each domain. We compare our results to recent style and photo transfer methods including Gatys et al.~\cite{gatys2016image}, WCT$^2$~\cite{yoo2019photorealistic}, STROTSS~\cite{kolkin2019style}, and our modified patch-based CycleGAN~\cite{zhu2017unpaired}. Our method can reproduce the texture of the reference photos while retaining structure of the input paintings. Our results are at 1k $\sim$ 1.5k resolution.}
 \lblfig{single1}
\end{figure}

\begin{figure}[t]
 \centering
  \includegraphics[angle=90, width=0.92\hsize]{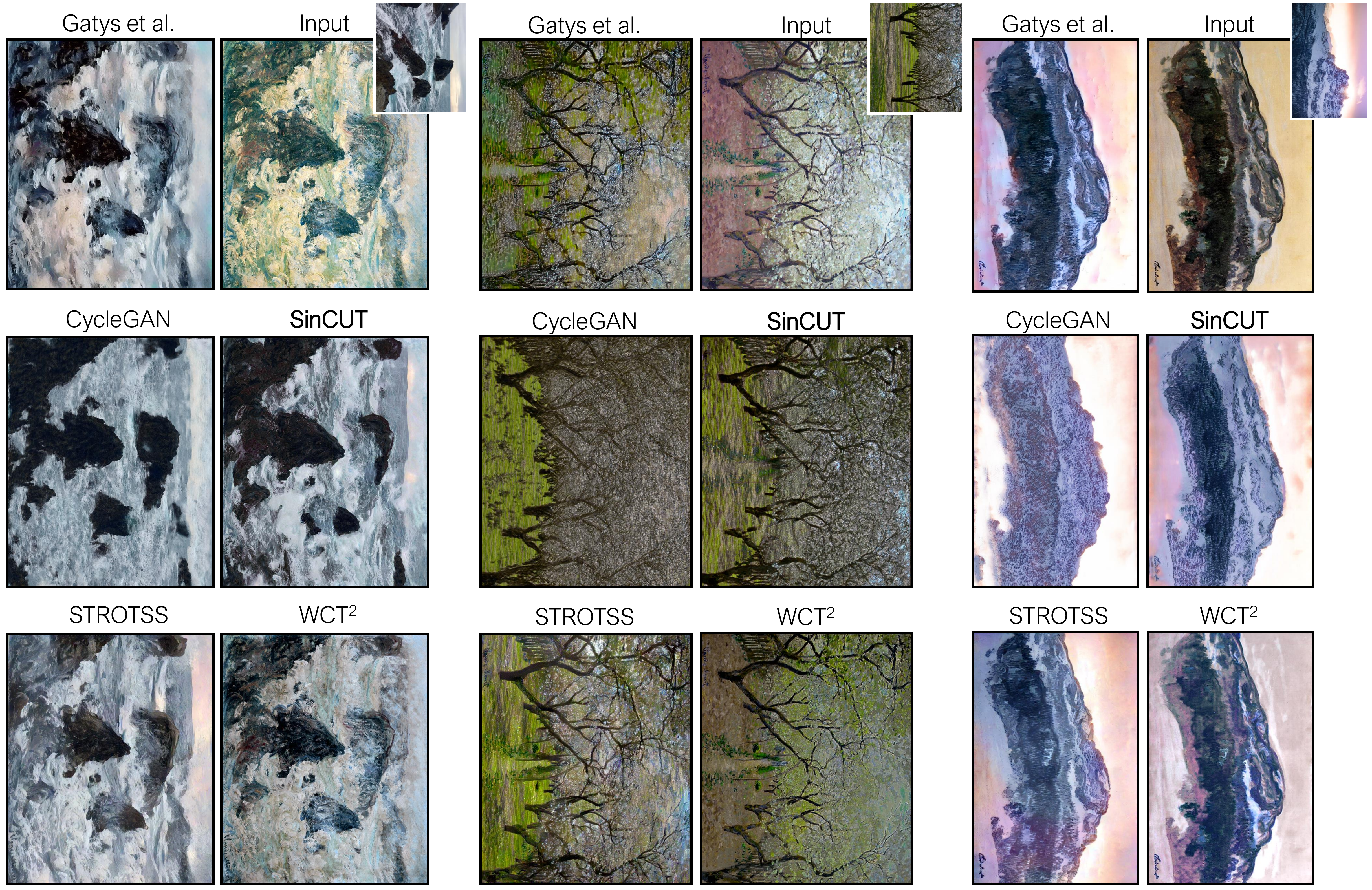}
  \vspace{-3mm}
 \caption{{\bf High-res painting to photo translation (II).} We transfer Monet's paintings to reference natural photos shown as insets at top-left corners. The training only requires a single image from each domain. We compare our results to recent style and photo transfer methods including Gatys et al.~\cite{gatys2016image}, WCT$^2$~\cite{yoo2019photorealistic}, STROTSS~\cite{kolkin2019style}, and our modified patch-based CycleGAN~\cite{zhu2017unpaired}. Our method can reproduce the texture of the reference photos while retaining structure of the input paintings. Our results are at 1k $\sim$ 1.5k resolution.}
 \lblfig{single2}
\end{figure}

\clearpage

\section{Unpaired Translation Details and Analysis}

\subsection{Training details}
\lblsec{app:training}
To show the effect of the proposed patch-based contrastive loss, we intentionally match the architecture and hyperparameter settings of CycleGAN, except the loss function. This includes the ResNet-based generator~\cite{johnson2016perceptual} with 9 residual blocks, PatchGAN discriminator~\cite{isola2017image}, Least Square GAN loss~\cite{mao2017least}, batch size of 1, and Adam optimizer~\cite{kingma2015adam} with learning rate 0.002.

Our full model CUT is trained up to 400 epochs, while the fast variant FastCUT is trained up to 200 epochs, following CycleGAN. Moreover, inspired by GcGAN~\cite{fu2019geometry}, FastCUT is trained with flip-equivariance augmentation, where the input image to the generator is horizontally flipped, and the output features are flipped back before computing the PatchNCE loss. Our encoder $\Genc$ is the first half of the CycleGAN generator~\cite{zhu2017unpaired}.
In order to calculate our multi-layer, patch-based contrastive loss, we extract features from 5 layers, which are RGB pixels, the first and second downsampling convolution, and the first and the fifth residual block. The layers we use correspond to receptive fields of sizes 1$\times$1, 9$\times$9, 15$\times$15, 35$\times$35, and 99$\times$99.  For each layer's features, we sample 256 random locations, and apply 2-layer MLP to acquire 256-dim final features. 
For our baseline model that uses MoCo-style memory bank~\cite{he2019momentum}, we follow the setting of MoCo, and used momentum value 0.999 with temperature 0.07. The size of the memory bank is 16384 per layer, and we enqueue 256 patches per image per iteration.

\subsection{Evaluation details}
\lblsec{app:evaluation}

We list the details of our evaluation protocol.

\myparagraph{Fr\'{e}chet Inception Distance (FID~\cite{heusel2017gans})} throughout this paper is computed by resizing the images to 299-by-299 using bilinear sampling of PyTorch framework, and then taking the activations of the last average pooling layer of a pretrained Inception V3~\cite{szegedy2016rethinking} using the weights provided by the TensorFlow framework. We use the default setting of \url{https://github.com/mseitzer/pytorch-fid}. All test set images are used for evaluation, unless noted otherwise.

\myparagraph{Semantic segmentation metrics on the Cityscapes dataset} are computed as follows. First, we trained a semantic segmentation network using the DRN-D-22~\cite{yu2017dilated} architecture. We used the recommended setting from \url{https://github.com/fyu/drn}, with batch size 32 and learning rate 0.01, for 250 epochs at 256x128 resolution. The output images of the 500 validation labels are resized to 256x128 using bicubic downsampling, passed to the trained DRN network, and compared against the ground truth labels downsampled to the same size using nearest-neighbor sampling.

\subsection{Pseudocode}
\lblsec{app:code}
Here we provide the pseudo-code of PatchNCE loss in the PyTorch style. Our code and models are available at our GitHub \href{https://github.com/taesungp/contrastive-unpaired-translation}{repo}.

\begin{minted}{python}

import torch
cross_entropy_loss = torch.nn.CrossEntropyLoss()

# Input: f_q (BxCxS) and sampled features from H(G_enc(x))
# Input: f_k (BxCxS) are sampled features from H(G_enc(G(x))
# Input: tau is the temperature used in NCE loss. 
# Output: PatchNCE loss
def PatchNCELoss(f_q, f_k, tau=0.07):
    # batch size, channel size, and number of sample locations
    B, C, S = f_q.shape
    
    # calculate v * v+: BxSx1
    l_pos = (f_k * f_q).sum(dim=1)[:, :, None]
    
    # calculate v * v-: BxSxS
    l_neg = torch.bmm(f_q.transpose(1, 2), f_k)
    
    # The diagonal entries are not negatives. Remove them.
    identity_matrix = torch.eye(S)[None, :, :]
    l_neg.masked_fill_(identity_matrix, -float('inf'))
    
    # calculate logits: (B)x(S)x(S+1)
    logits = torch.cat((l_pos, l_neg), dim=2) / tau
    
    # return NCE loss
    predictions = logits.flatten(0, 1)
    targets = torch.zeros(B * S, dtype=torch.long)
    return cross_entropy_loss(predictions, targets)

\end{minted}

\begin{figure}
  \centering
  \begin{minipage}{1.0\linewidth}
  \includegraphics[width=\linewidth]{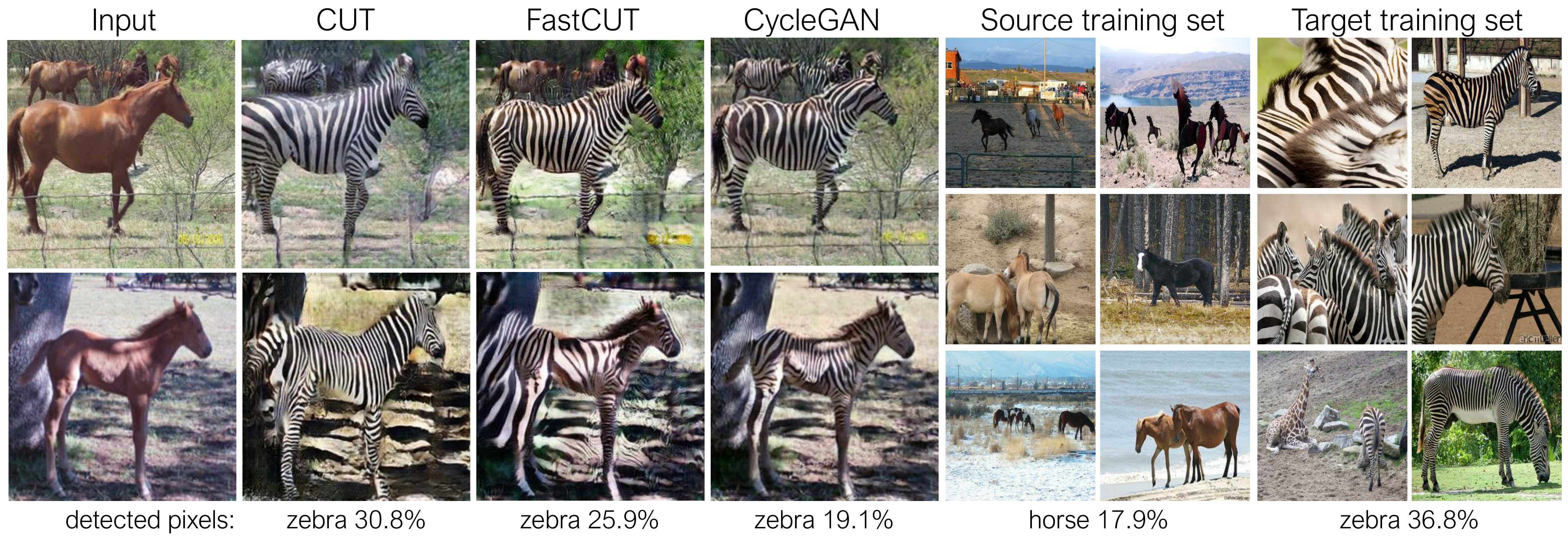} 
  \end{minipage} \\
  
  \caption{\small \textbf{Distribution matching.} We measure the percentage of pixels belonging to the horse/zebra bodies, using a pre-trained semantic segmentation model. We find a distribution mismatch between sizes of horses and zebras images -- zebras usually appear larger (36.8\% vs. 17.9\%). Our full method CUT has the flexibility to enlarge the horses, as a means of better matching of the training statistics than CycleGAN~\cite{zhu2017unpaired}. Our faster variant FastCUT, trained with a higher PatchNCE loss ($\lambda_{\Xset}=10$) and flip-equivariance augmentation,  behaves more conservatively like CycleGAN. 
    }
  \lblfig{distribution}
\end{figure}

\subsection{Distribution matching}
\lblsec{app:distribution}

\lblsec{distribution_matching}
In \reffig{distribution}, we show an interesting phenomenon of our method, caused by the training set imbalance of the horse$\rightarrow$zebra set. We use an off-the-shelf DeepLab model~\cite{chen2018deeplab} trained on COCO-Stuff~\cite{caesar2018coco}, to measure the percentage of pixels that belong to horses and zebras\footnote{Pretrained model from \url{https://github.com/kazuto1011/deeplab-pytorch}}. The training set exhibits dataset bias~\cite{torralba2011unbiased}. On average, zebras appear in more close-up pictures than horses and take up about twice the number of pixels ($37\%$ vs $18\%$). To perfectly satisfy the discriminator, a translation model should attempt to match the statistics of the training set. Our method allows the flexibility for the horses to change the size, and the percentage of output zebra pixels ($31\%$) better matches the training distribution ($37\%$) than the CycleGAN baseline ($19\%$). On the other hand, our fast variant \textit{FastCUT} uses a larger weight ($\lambda_{\Xset} = 10$) on the Patch NCE loss and flip-equivariance augmentation, and hence behaves more conservatively and more similar to CycleGAN. The strong distribution matching capacity has pros and cons. For certain applications,  it can create introduce undesired changes (e.g., zebra patterns on the background for horse$\rightarrow$zebra). On the other hand, it can enable dramatic geometric changes for applications such as Cat$\rightarrow$Dog. 

\subsection{Additional Ablation studies}
\lblsec{app:ablation}
In the paper, we mainly discussed the impact of loss functions and the number of patches on the final performance. Here we present additional ablation studies on more subtle design choices. We run all the variants on horse2zebra datasets~\cite{zhu2017unpaired}. The FID of our original model is {\bf 46.6}. We compare it to the following two variants of our model:  
\begin{itemize}[leftmargin=*,noitemsep,topsep=0pt,label=\textbullet]
\item Ours without weight sharing for the encoder $\Genc$ and MLP projection network $\proj$:
for this variant, when computing features $\{\z_l\}_L=\{\proj_l(\Genc^l(\x))\}_L$, we use two separate encoders and MLP networks for embedding input images (e.g., horse) and the generated images (e.g., zebras) to feature space. They do not share any weights. The FID of this variant is {\bf 50.5}, worse than our method. This shows that weight sharing  helps stabilize training while reducing the number of parameters in our model.  
\item Ours without updating the decoder $\Gdec$ using \textit{PatchNCE} loss: in this variant, we exclude the gradient propagation of the decoder $\Gdec$ regarding \textit{PatchNCE} loss $\Lpatchnce$. In other words, the decoder $\Gdec$ only gets updated through the adversarial loss $\Lgan$. The FID of this variant is {\bf 444.2}, and the results contain severe artifacts. This shows that our $\Lpatchnce$ not only helps learn the encoder $\Genc$, as done in previous unsupervised feature learning methods~\cite{he2019momentum}, but also learns a better decoder $\Gdec$ together with the GAN loss. Intuitively, if the generated result has many artifacts and is far from realistic, it would be difficult for the encoder to find correspondences between the input and output, producing a large \textit{PatchNCE} loss. 
\end{itemize}

\section{Changelog}
\myparagraph{v1} Initial preprint release (ECCV 2020)

\myparagraph{v2 and v3} (1) Fix typos in \refeq{nce_int} and \refeq{nce_ext}. (2) Add additional related work.
\end{subappendices}

\end{document}